\documentclass[lettersize,journal]{IEEEtran}
\usepackage{amsmath,amsfonts}
\usepackage{algorithmic}
\usepackage{algorithm}
\usepackage{array}
\usepackage{textcomp}
\usepackage{stfloats}
\usepackage{verbatim}
\usepackage{graphicx}
\usepackage{cite}
\usepackage{subfigure}
\usepackage{subcaption}

\usepackage{xurl}
\usepackage{tabularx}
\usepackage{booktabs}
\usepackage{hyperref}
\usepackage{cleveref}
\begin{document}

\title{Drone Detection using Deep Neural Networks Trained on Pure Synthetic Data}

\author{Mariusz Wisniewski\thanks{Mariusz Wisniewski is with Cranfield University, College Road, Cranfield, Bedfordshire, MK43 0AL. (e-mail: m.wisniewski@cranfield.ac.uk)}, Zeeshan A. Rana\thanks{Zeeshan A. Rana is with Cranfield University, College Road, Cranfield, Bedfordshire, MK43 0AL and Prince Mohammad Bin Fahd University.}, Ivan Petrunin\thanks{Ivan Petrunin is with Cranfield University, College Road, Cranfield, Bedfordshire, MK43 0AL. }, Alan Holt\thanks{Alan Holt is with Department for Transport.}, Stephen Harman\thanks{Stephen Harman is with Thales UK.}}%

\maketitle

\begin{abstract}
Drone detection has benefited from improvements in deep neural networks, but like many other applications, suffers from the availability of accurate data for training. Synthetic data provides a potential for low-cost data generation and has been shown to improve data availability and quality. However, models trained on synthetic datasets need to prove their ability to perform on real-world data, known as the problem of sim-to-real transferability.
Here, we present a drone detection Faster-RCNN model trained on a purely synthetic dataset that transfers to real-world data. 
We found that it achieves an $AP_{50}$ of 97.0\% when evaluated on the MAV-Vid - a real dataset of flying drones - compared with 97.8\% for an equivalent model trained on real-world data. Our results show that using synthetic data for drone detection has the potential to reduce data collection costs and improve labelling quality. 
These findings could be a starting point for more elaborate synthetic drone datasets. For example, realistic recreations of specific scenarios could de-risk the dataset generation of safety-critical applications such as the detection of drones at airports. Further, synthetic data may enable reliable drone detection systems, which could benefit other areas, such as unmanned traffic management systems. The code\footnote{The code is available \url{https://github.com/mazqtpopx/cranfield-synthetic-drone-detection}} and datasets\footnote{The datasets are available \url{https://huggingface.co/datasets/mazqtpopx/cranfield-synthetic-drone-detection} } are available.
\end{abstract}


\section{Introduction}
\IEEEPARstart{D}{rone} intrusions pose a risk to airports. In 2018, a reported drone sighting disrupted the operations at the Gatwick airport, UK, for 3 days. Since then, drones have been used for other illicit purposes, such as smuggling drugs into prisons \cite{IncreaseUseDrones2023, DrugsWeaponsSmuggled2022}. Although a serious disruption, such as the one in Gatwick, has not occurred since, reports of drone sightings at UK airports persist \cite{HeathrowAirportDrone2019,skopelitiFlightsDivertedEast2022, DublinAirportFlights2023}. 
To prevent this, the UK Government presented a counter-unmanned aircraft strategy \cite{UKCounterUnmannedAircraft}. It sets the detection of drones as a key objective.

On the other hand, UAS traffic management (UTM) is a relatively new area of research. It is concerned with integrating drones and air taxis with the current air traffic management (ATM). 
For UTM to work in urban environments, camera systems will likely have to work alongside RF/radar sensors due to interference from buildings. 
Accurate detection systems will enable the use of authorized drones for approved unmanned operations, whilst preventing unauthorized drones from flying in excluded parts of the airspace.

Drones can be detected using different sensors such as radar, radio frequency (RF), acoustic, and electro-optical/infrared \cite{lykouDefendingAirportsUAS2020}. State-of-the-art detection methods typically fuse some of these sensors. Our research focuses on drone detection using a visible spectrum camera.

Recent advances in neural networks (NNs) allow for much more accurate detection, tracking, and classification of drones in images or video feeds. However, accurate data for training the NN models remains difficult to obtain. In this article, we explore the use of structured domain randomization (SDR) within a simulated environment to generate a synthetic dataset of drones. We test whether a synthetic dataset generated using this method is generalizable to real-world datasets, by comparing the results to an established benchmark.

\subsection{Literature}
We focus on the optical detection of drones using NNs trained on synthetic data. Although object detection in images has been extensively studied, drone detection poses unique challenges, usually caused by environmental, or systematic issues. 
For example, weather effects impact the image, the drones can be small in the video frame making them hard to detect or classify,  and the movement of UAVs is complicated to predict. Classification of the objects in the sky such as drones and birds is also a key issue. 
We investigate the literature on NNs, the use of synthetic data for training NNs, applications of NNs for drone detection, and finally, the use of synthetic data for drone detection.

\subsubsection{DNN advances and architectures}\label{subsection:introduction:dnn}
Recent advances in deep neural network (DNN) architectures (e.g. AlexNet \cite{krizhevskyImageNetClassificationDeep2017}) became the state-of-the-art method for object recognition. They were later expanded to detect object bounding boxes, segment objects, and track them across videos. 
Object detection architectures comprise of single-stage and two-stage detectors. Examples of one-stage detectors are YOLO \cite{redmonYouOnlyLook2016}, \cite{redmonYOLO9000BetterFaster2017} and SSD \cite{liuSSDSingleShot2016}. Examples of two-stage detectors are Fast-RCNN \cite{girshickFastRCNN2015}, Faster-RCNN \cite{renFasterRCNNRealTime2016}. 
One-stage detectors do the detection/classification directly from the extracted features, without a separate region proposal. 
Two-stage detectors generate region proposals (i.e. regions of the image where it thinks an object might exist) and then classify each of the proposed regions. 
In theory, two-stage detectors should produce better accuracy at a higher computational cost than one-stage detectors. However, in practice, they tend to vary for different applications and are sensitive to training parameters. An example of this is Isaac-Medina et al. \cite{isaac-medinaUnmannedAerialVehicle2021}, where the results of different architectures vary across different datasets. 
More recently, transformers \cite{vaswaniAttentionAllYou2017} which use an attention mechanism and an encoder-decoder architecture, showed improvements in performance over traditional RNNs and CNNs as they are more parallelizable.  This led to detection architectures such as DETR \cite{carionEndtoEndObjectDetection2020}. DETR matches Faster R-CNN in terms of performance on the COCO object detection dataset.

\subsubsection{Synthetic Data, Sim-to-Real, and Domain Randomization} \label{subsection:synthetic_data}
Accurate data for training neural networks is expensive and may be hard to obtain. Creating real-life images or videos has a cost associated with it, which varies depending on the application.
After the data is recorded, it needs to be manually labelled with the ground truth by a human. 
Synthetic data offers an automated alternative. By using 3D models and rendering software, synthetic 
images along with accurate ground truth labels can be generated. 
This presents a research question: \textit{can images rendered by 3D software be used to train NN models and transferred to real-world applications? }

The use of synthetic data comes with advantages and disadvantages. The primary advantage is the potential reduction in the time cost required to generate the dataset. An algorithm can generate a very large dataset with no human input required. The primary disadvantage is that NN models trained on synthetic data are hard to transfer to real-world problems. Hence, the primary objective of synthetic data research is bridging the sim-to-real gap - the transfer of a NN model trained on synthetic data to perform equivalently on real-world data. 

However, machine learning algorithms used to train neural networks rely on the training and evaluation datasets coming from the same source distribution. We are fundamentally breaking this assumption by using synthetic data for training and evaluating on real-world data. This is referred to as the out-of-distribution generalization \cite{liuOutOfDistributionGeneralizationSurvey2023}. 

There exist multiple strategies for achieving out-of-distribution generalization such as transfer learning \cite{zhuangComprehensiveSurveyTransfer2020}, domain adaptation \cite{csurkaDomainAdaptationVisual2017}, and domain generalization \cite{zhouDomainGeneralizationSurvey2022}. Each of these relies on different assumptions between the training data, access to test data, and training conditions. To achieve sim-to-real transfer, domain generalization is the most relevant to our problem, as we assume that we cannot see the real-world test data during training. For the problem of sim-to-real for images, the most popular method of domain generalization is domain randomization. 
 
\textbf{Domain randomization} (DR) is a method of introducing model generalizability. This is done by randomizing parameters used to synthesize the source distribution, with the goal of making the data distribution so wide and varied that the model trained on this dataset is transferable to the target distribution.
Lighting, textures, poses, and other parameters are randomized. 
Popular approaches aim to make the scenes as unrealistic as possible. Tobin at al. \cite{tobinDomainRandomizationTransferring2017} is an example, which employs DR by randomizing the textures of the objects to make the network invariant to changes in scene conditions. 
Trembley et al. \cite{tremblayTrainingDeepNetworks2018} use DR to bridge the sim-to-reality gap for object detection of cars for an autonomous driving application. Their approach employs DR by inputting random shapes, textures, and cars in unrealistic positions, with the aim of generating more variety to focus the neural networks on the structure of the object of interest. 
Prakash et al. \cite{prakashStructuredDomainRandomization2019} propose the use of structured domain randomization (SDR) - a variant of domain randomization, which takes into account the context of the scene. Hence instead of generating random scenes with unrealistic textures and objects, it aims to generate realistic scenes, while randomizing other parameters in a structured way. They find that SDR provides performance improvements over DR. 
Borrego et al. \cite{borregoApplyingDomainRandomization2018} propose the use of synthetic DR datasets as an alternative to pre-trained networks.  
Alghonaim \& Johns \cite{alghonaimBenchmarkingDomainRandomisation2021} investigate the parameters used to randomize the simulation such as camera position, target colour, and lighting variations to find the most important parameters for sim-to-real transfer. They found that more high-quality realistic images improved the performance of sim-to-real. This comes in contrast with some of the earlier works (Tobin et al. \cite{tobinDomainRandomizationTransferring2017} and Trembley et al. \cite{tremblayTrainingDeepNetworks2018}) which focused on unrealistic, low-quality generated images.  They also found that the randomization of distractor objects and background textures was important for generalising to new environments. 
Hinterstoisser et al. \cite{hinterstoisserPreTrainedImageFeatures2017} propose a 'trick' to freeze the layers responsible for feature extraction of a NN trained on real-world images, and train only the remaining layers on synthetic images. They show that their approach works well on architectures such as Faster-RCNN, and Mask-RCNN.

Hence, there appear two schools of thought. DR produces unrealistic images with the aim of expanding the domain to allow the NN to better learn the features in settings that might not necessarily be used during test time. SDR produces realistic images with the aim of making the training data similar to the real-world data used at test time. 

\subsection{Visual Drone Detection}
Visual drone detection is the process of finding the position of a drone in images or video feeds produced by visible-spectrum cameras. It has become a popular area of research, largely driven by the recent advances in machine learning. Most approaches
\cite{aydinDroneDetectionUsing2023, samadzadeganDetectionRecognitionDrones2022a,fanObjectDetectionAlgorithm2022a, dadrassjavanModifiedYOLOv4Deep2022}
typically use some of the detection architectures described in section \ref{subsection:introduction:dnn}, or a modification of the network for drone detection purposes, trained and tested on one of the publicly available drone datasets, or on a private dataset.

Isaac-Medina et al. \cite{isaac-medinaUnmannedAerialVehicle2021} present a benchmark of multiple DNN algorithms (SSD, YOLOv3, Faster R-CNN, and DETR) on publicly available datasets of flying drones (MAV-VID, Drone-vs-Bird, Anti-UAV). They present two benchmarks: the accuracy of detection, and the accuracy of tracking algorithms.
They present the results of each of the models trained and tested on each of the datasets. Note that they create three distinct sets of weights for each of the datasets.   
We were able to independently reproduce the test metrics shown by Isaac-Medina et al. based on their published model weights for the Faster-RCNN model (although, we did not attempt to reproduce the training of the models). 

Freudenmann et al. \cite{freudenmannExploitationDataAugmentation2021} explore the effects of data augmentation strategies on the performance of their NN models. They set up their experiments by training a Faster R-CNN model on the Drone vs Bird dataset. They then test their NN model on other datasets: a subset of the Drone vs Bird dataset, Anti-UAV dataset, background test dataset (images of drones scraped from the internet), Disturbing Objects dataset (to find false positives), and New UAV Types (unseen UAV images scraped from the internet).  
They find that by test a model trained on the DvB dataset, on the Anti-UAV dataset, the accuracy drops significantly. The authors propose the cause to be the mosaics and artefacts found in the Anti-UAV dataset. To reduce this effect, the authors propose the use of mosaic + standard dataset transforms. This increases the mAP to 78.7\% from  54.2\% with no augmentations. The augmentations appear to train the model to reduce the false positive rate. Still, this falls short of the results presented by Isaac-Medina et al. where the Faster R-CNN model trained on the Anti-UAV dataset and tested on the subset of the Anti-UAV dataset achieves 97.7\%. 
This highlights the challenge of transferring models trained on one dataset to other datasets - a drone detection model trained on one dataset is not guaranteed to perform equivalently on another dataset.

Mediavilla et al. \cite{mediavillaDetectingAerialObjects2021} present a benchmark of Swin, YOLOv4, CenterNet, CenterNet2, and Faster-RCNN architectures on CACHOW - a helicopter dataset - and Drone vs Bird datasets. However, the results are not directly comparable with Isaac-Medina et al. as the test configurations are different. 

Other approaches consider the temporal dimension. Thai et al. \cite{thaiSmallFlyingObject2023} present a spatio-temporal NN design, in which multiple frames of the video feed are stacked together before being input into the NN. They show that this approach is more accurate, but comes with a higher computational cost as more data needs to be processed. 
Craye and Ardjoune \cite{crayeSpatioTemporalSemanticSegmentation2019} use U-Net to segment the position of the drone, classify it, and then use a spatio-temporal filter. Sangam et al. \cite{sangamTransVisDroneSpatioTemporalTransformer2023} present a spatio-temporal transformer for drone-to-drone detection.

\subsection{Synthetic Data for Drone Detection}
Synthetic data approaches have been attempted in the field of drone detection. One of the first attempts that we were able to find is Rozantsev et al. \cite{rozantsevRenderingSyntheticImages2015} which uses domain randomization techniques (without referring to them as such - the paper was published before domain randomization became a common term in literature), such as randomizing the position of the drones, varying the motion blur, adding random noise, and randomizing object material properties. They also repeat this process for aircraft and cars. 

Marez et al. \cite{marezUAVDetectionDataset2020} create a synthetic dataset of UAVs using DR and test it on the Drone vs Bird dataset. They create a baseline model (trained only on DvB) and compare the models trained only on the DR datasets, and models pre-trained on the DR datasets and finetuned on real-world data. They find that DR-only datasets perform the worst, while the finetuned models perform better but still fall short of the baseline. This highlights the challenge of transferring synthetically trained models to real-world datasets.
Peng et al. \cite{pengUsingImagesRendered2018} use 3D models of drones and random HDRIs to generate a photorealistic synthetic dataset, and train a Faster-RCNN model. It is tested on a real-life dataset of images downloaded from the internet. They find that using a pre-trained model improves their performance significantly, and using occluded drone images improves their performance by ~1\%. However, their test dataset of images downloaded from the internet might not be representative of a real-world security scenario. We were also unable to find a copy of their test dataset online which makes it hard to directly compare the results to. 
Dieter et al. \cite{dieterQuantifyingSimulationReality2023} attempt to bridge the sim-to-real gap by generating a synthetic dataset in Unreal Engine. They test their models on synthetic and real data. Overall, they find that, generally, models trained on synthetic data fall short in terms of the performance of the models trained on real-world data. However, by using a small share of real-world data, they improve the results and get close to the results produced by using real-world data only.  
DronePose \cite{scholesDronePoseIdentificationSegmentation2021} uses 3D models of a DJI Mavic, and a DJI Inspire, to create a synthetic dataset using Unreal Engine. They also create a neural network based on a U-Net architecture, to segment parts of the drone, find the orientation, and identify the drone model. 
The use of GANs to generate a synthetic dataset was attempted by Li et al.  \cite{liScarceDataDriven2021}. 
This is done to better understand the deep feature space of how drones are represented by neural networks. 
Further, they use Topological Data Analysis to acquire missing data. 
This study looks at something that not many other studies consider - how neural networks learn to represent drones within the latent space. 
Carrio et al. \cite{carrioDroneDetectionUsing2018} use synthetic images to predict drones using depth maps for the application of obstacle detection of avoidance. By using a synthetic environment, they are able to create accurate ground truth depth maps, which they later use to train a NN.

\subsection{Outro - Contribution and Outline}
This work is largely a continuation of Wisniewski et al. \cite{wisniewskiDroneModelClassification2022, wisniewskiDroneModelIdentification2021}. It expands the method of synthetic data generation to object detectors, which can detect the position of the drone in each of the frames. Isaac-Medina et al. created a UAV detection benchmark, in which they compare common object detection architectures. This benchmark is used as a baseline to compare the results from our models trained on synthetic data.

We present a method of systematically generating a synthetic dataset of drones alongside pixel-accurate ground truth segmentation masks using a form of structured domain randomization. This presents an advantage over real-world datasets which are costly to annotate and may not be as accurate. Further, synthetic datasets allow flexibility over standard approaches
such as being able to add new drones to the dataset using only a 3D model. They also allow for the creation of complex scenarios that may be hard to record in real life such as forest fires, or drones flying in stormy conditions. Generally, drones should not be flown in windy conditions in real life and this might make creating a research study case hard due to safety concerns.

The challenge with this approach is that the synthetic dataset may not be representative of the real-world data, and a model trained on this dataset might not translate to real-world datasets - this is the sim-to-real problem of transferring domains, as explored in section \ref{subsection:synthetic_data}. 
We present and compare multiple datasets which are generated by processes of SDR and DR. 
The reasons for focusing on the dataset are:
\begin{itemize}
    \item The publicly available drone datasets vary in quality. Because they use human-labeled ground truth boxes, these are imperfect at times. 
    They are also limited in that they do not have the segmentation ground truth. 
    \item Some of the previously publicly available datasets are no longer available.
    \item New real-world datasets are costly to produce.
\end{itemize}

These items make researching drone detection and comparing results between studies challenging. We address these points by releasing our synthetic dataset. 
The contributions of this publication are:
\begin{itemize}
    \item Multiple synthetic datasets with pixel-accurate segmentation masks, generated by the process of structured domain randomization.
    \item A study into the ability of the neural network model trained on synthetic data to generalize to real-world drone detection scenarios. This differs from SOTA in the review because the synthetic drone detection applications mixed real-world data with their synthetic data for the training to achieve sim-to-real. They also didn't use open-access datasets such as Drone-vs-Bird or Anti-UAV. 
    \item Study of different domain randomization styles.
    \item Generalizable Faster R-CNN model weights for drone detection; trained purely on synthetic images, tested on multiple real-life datasets.
\end{itemize}

Through our literature review, we have identified several faults within some of the research in the drone detection area that we attempted to remedy during our study:
\begin{itemize}
    \item \textbf{Use of private datasets}. Using private datasets makes it impossible for other researchers to reproduce results. To remedy this, we open-source our dataset, and we use publicly available datasets for testing.
    \item \textbf{Lack of a common baseline to compare the results to}. This is a common problem which makes it hard to directly compare the results of one method over another. To remedy this, we use the results of the Isaac-Medina et al. benchmark to compare our results. 
    \item \textbf{Repeatability of results}. Many publications present the results as a single value. However, due to the randomness of neural network training, retraining yields different results. To remedy this, we present our results as a mean with a 95 per cent confidence interval. 
    \item \textbf{Reproducibility of results}. A lot of research in this area is hard to reproduce. To remedy this, we open-source\footnote{Datasets \url{https://huggingface.co/datasets/mazqtpopx/cranfield-synthetic-drone-detection}, and source code \url{https://github.com/mazqtpopx/cranfield-synthetic-drone-detection}} our synthetic dataset, as well as the code used for training and testing of the NN models. 
\end{itemize}

We believe that this is one of the first attempts to study the use of structured domain randomization for the generation of synthetic drone datasets.
Further, we perform a study comparing different domain randomization techniques to find out how much they influence drone detection. Although we have found attempts in the literature to use domain randomization techniques for drone detection, we were unable to find systematic studies to quantify which aspects of domain randomization improve the results of drone detection. 

To the best of our knowledge, previous attempts to test the sim-to-real transferability used private real-world datasets. Instead, we directly compare our results to the benchmark presented by Isaac-Medina et al. \cite{isaac-medinaUnmannedAerialVehicle2021} which uses publicly available datasets. 

We present a successful sim-to-real transfer for the application of drone detection. We prove that our NN model trained on a purely synthetic dataset successfully transfers to the MAV-Vid dataset. Our model achieves a mean of 97.0\% compared with the Isaac-Medina et al. model, which was trained on the MAV-Vid dataset, of 97.8\%. 

This paper is divided into five sections. In section \ref{section:Methodology} we explain the methodology. This includes explaining the rendering process used to generate the datasets. We explain different styles of datasets generated using different domain randomization styles. We describe the NN training process, we explore the use of data augmentations, in particular noise and JPEG compression, and finally, we explain the testing procedure, to measure the accuracy of sim-to-real transfer. In section \ref{section:Results} we explain the results. We perform tests on camera bounds randomization, effects of data augmentations, and variation in domain randomization styles, and lastly, we compare our results to the literature. In section \ref{section:Conclusion} we conclude the findings. In section \ref{section:Further_Work} we discuss further work. 

\section{Methodology}\label{section:Methodology}
\begin{figure}
\centering
\includegraphics[width=1.0\linewidth]{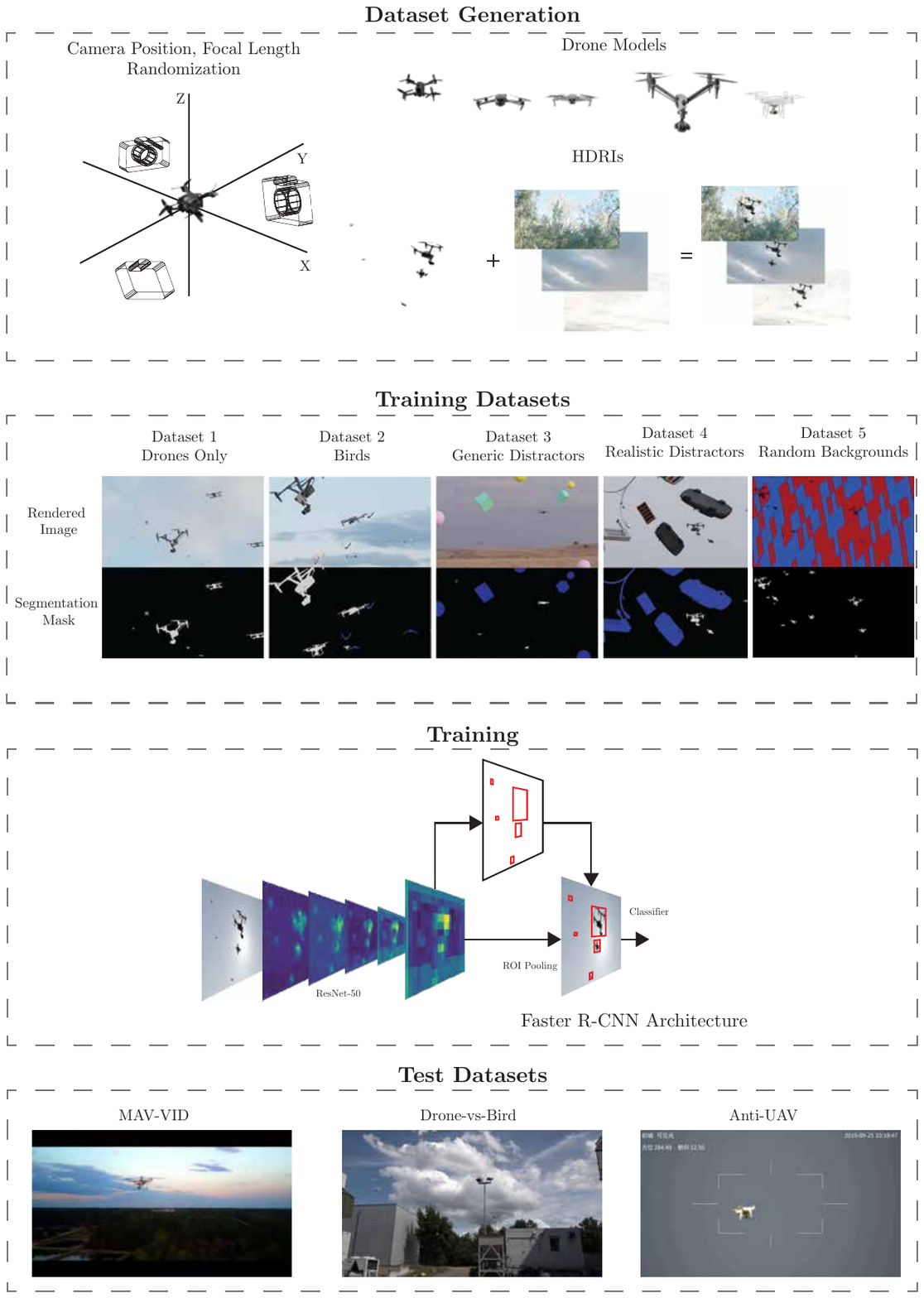}
\caption{The experimental process is split into 3 parts: synthetic dataset generation, training of neural networks (NNs), and testing. During the synthetic dataset generation, the position of the camera and HDRIs (background and lighting setup) are randomized.  4 distinct synthetic datasets are created: drones only, drones and birds, drones and generic distractors, drones and detailed distractors.  A Faster-RCNN model is trained on the synthetic datasets. The Faster-RCNN model is then tested on each of the real-world datasets: MAV-VID, Drone-vs-Bird, and Anti-UAV.  \label{fig:methodology:experimental_process}}
\end{figure}

This section describes the 3 stages of our experimental process: synthetic dataset generation (section \ref{section:methodology:synthetic_dataset_generation}), training of neural networks (NNs) (section \ref{section:methodology:neural_network_training}), and the testing of the NN models on real-world datasets (section \ref{section:methodology:neural_network_testing}). The experimental process broken up into these stages is shown in figure \ref{fig:methodology:experimental_process}. 

\subsection{Synthetic Dataset Generation using Structured Domain Randomization}\label{section:methodology:synthetic_dataset_generation}
This section describes the methodology used to create synthetic datasets using Blender \cite{blenderfoundationBlender3DModelling} for the purpose of drone detection. This is done by the process of Structured Domain Randomization (SDR) \cite{prakashStructuredDomainRandomization2019}. 
Blender allows the rendering of photorealistic images using a ray-trace-based Cycles engine. It also allows for programmatic randomization of simulation parameters, and the creation of pixel-level accurate ground truth segmentation masks. 
The primary advantage of this approach is that the labelling of ground truth is automated - compared with real-life approaches, where manual labelling is required. 
We continue this section by describing the process of SDR used to generate the datasets and the multiple datasets that are generated using different styles. 

SDR is a technique for generating random synthetic images. It differs from domain randomization (DR) in that it aims to generate scenes that have a realistic context. 
DR aims to expand the domain used to train the NN model outside of the operating domain to improve the performance of the model within the operating domain. This is achieved by implementing unrealistic textures, backgrounds, and contexts. SDR differs in that it aims to recreate the operating domain within the simulation and uses realistic textures, backgrounds, and contexts, but continues to randomize global parameters (such as lighting, camera poses, and layout of scenarios). 

Hence, we will refer to realistic datasets as SDR, and unrealistic datasets as DR. We will refer to a drone with realistic textures in a realistic context as SDR, and we will refer to a flying cube object with a random texture as DR. Although our drone dataset is generated using the SDR technique, we also attempt the use of DR methods by creating unrealistic distractors in datasets with generic and realistic distractors (these can be seen in Figure \ref{fig:methodology:synth_drones}).

\subsubsection{Rendering Process}\label{section:methodology:rendering_process}
We use a Python script within Blender to generate the datasets. 
We use 5 different 3D models of drones: DJI Phantom, DJI Inspire, 2 DJI Mavics, and a DJI FPV. We model the drones as a swarm travelling in a path from one point to another. 
The drones are animated for 300 frames. After the 300 frames, a new path is generated and the HDRI (background and lighting) is changed. 
Every frame, the position of the camera and the focal length is randomized. The camera is influenced to look towards the drones. 
The focus point of the camera is set to a very large distance. This is done to model security cameras, which are typically focused on infinity. A limitation of Blender is that it does not model the function to focus on infinity within its camera model, so we get around this limitation by setting the focus distance to a very large value. 
The illustrations of the DJI drone models and the overall dataset generation process are shown at the top of figure \ref{fig:methodology:experimental_process}).

\subsubsection{Variations in Synthetic Datasets}\label{section:methodology:variation_in_datasets}
We create a few variations of the datasets based on different methods from the literature to test if they improve the performance of the detection model. The variations that we create are drones with birds, drones with generic distractors, drones with realistic distractors, and drones with unrealistic backgrounds. 

\begin{figure}
    \subfigure[Drones Only]{        
        \includegraphics[trim={840px 0 0 0},clip,width=.24\linewidth]{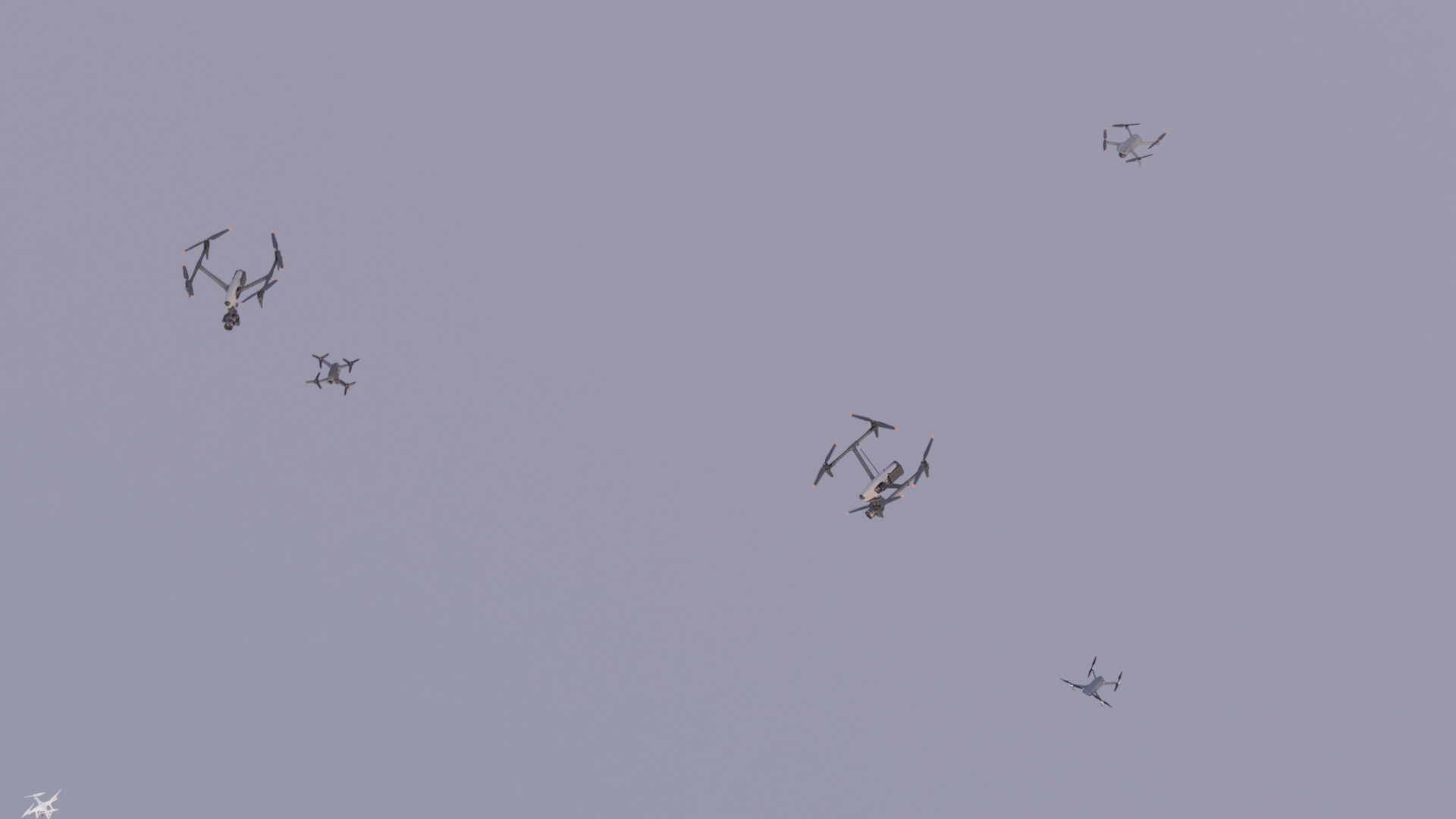}\hfill
        \includegraphics[trim={840px 0 0 0},clip,width=.24\linewidth]{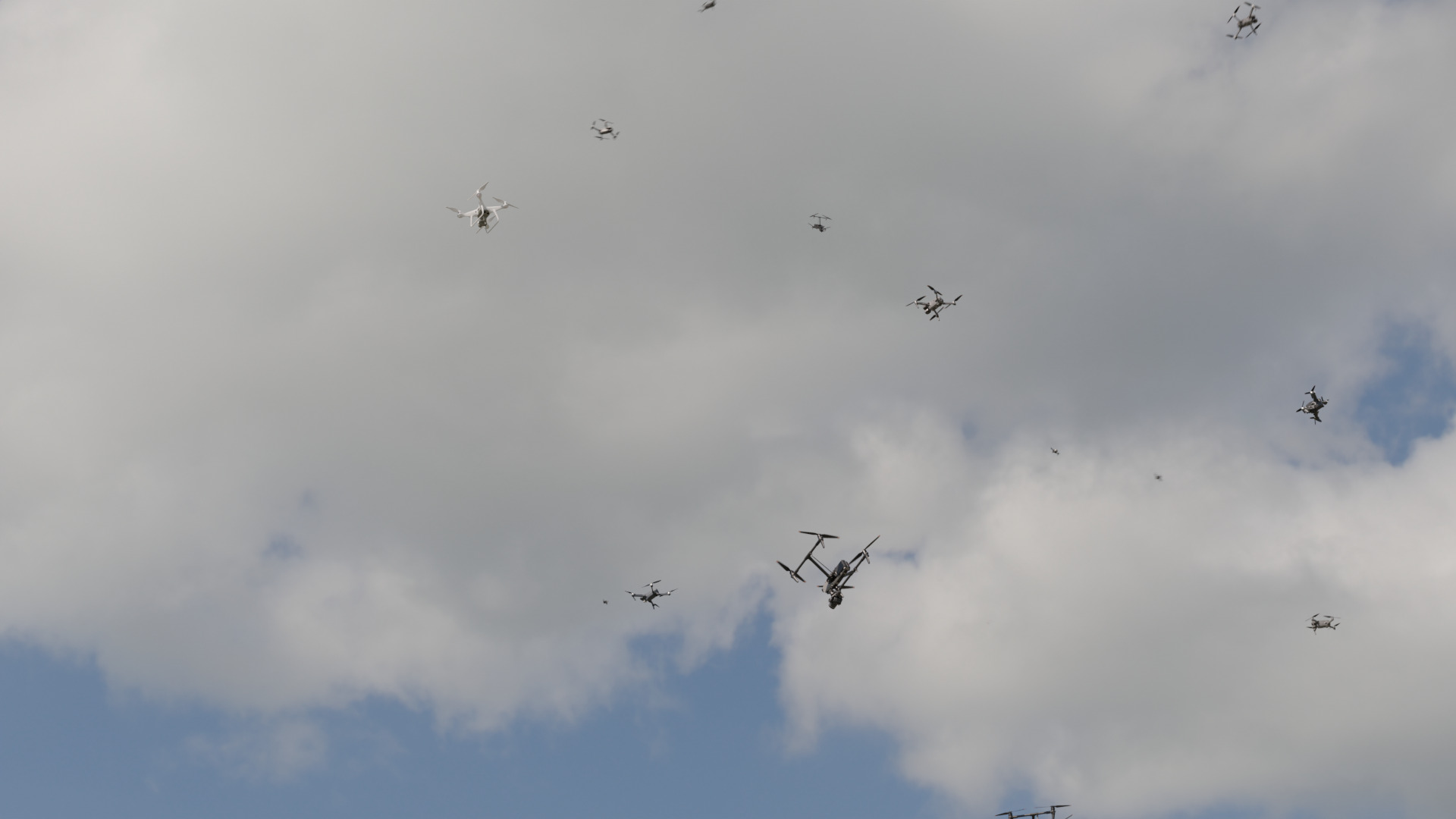}\hfill
        \includegraphics[trim={0 0 840px 0},clip,width=.24\linewidth]{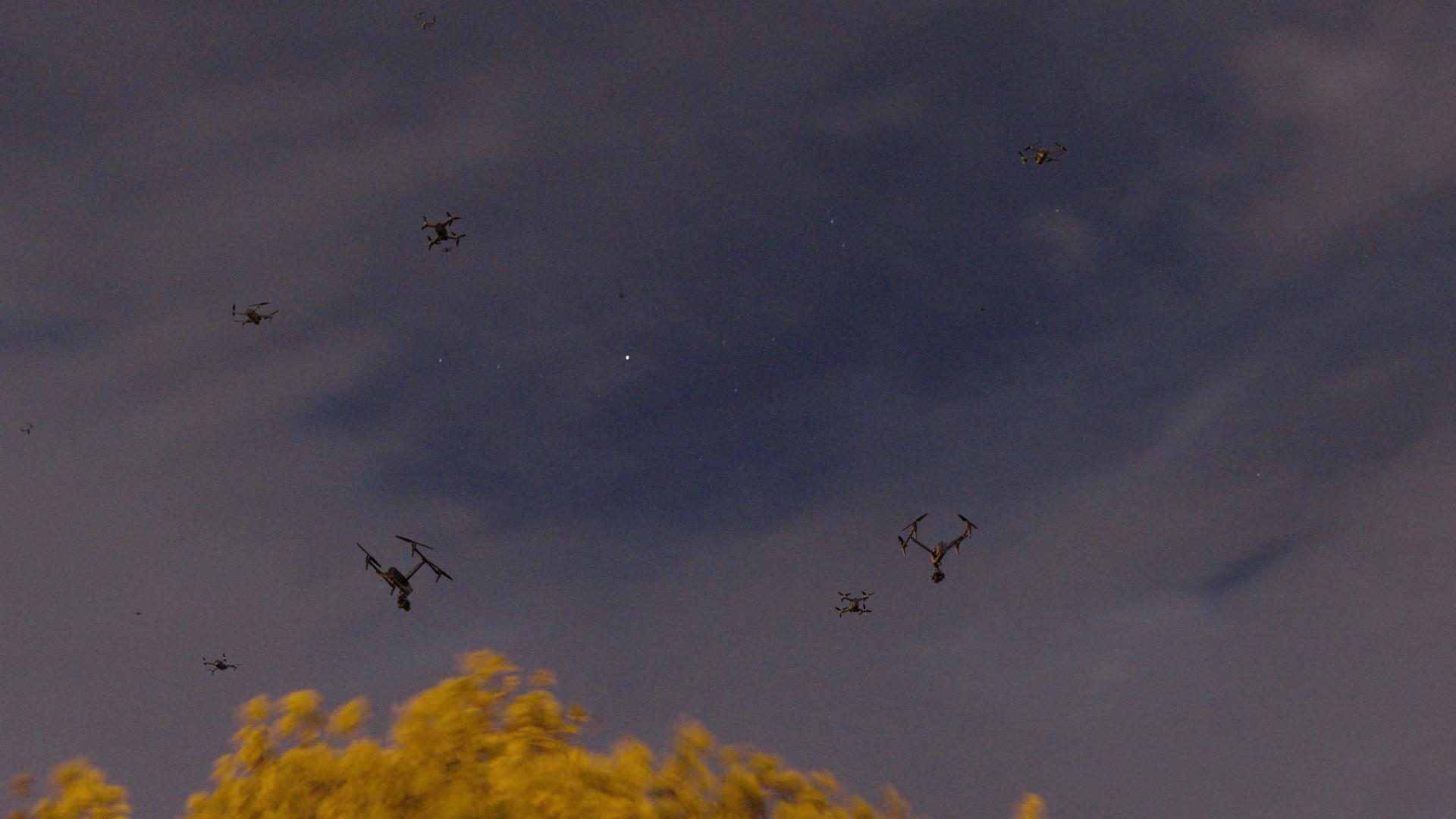}\hfill
        \includegraphics[trim={840px 0 0 0},clip,width=.24\linewidth]{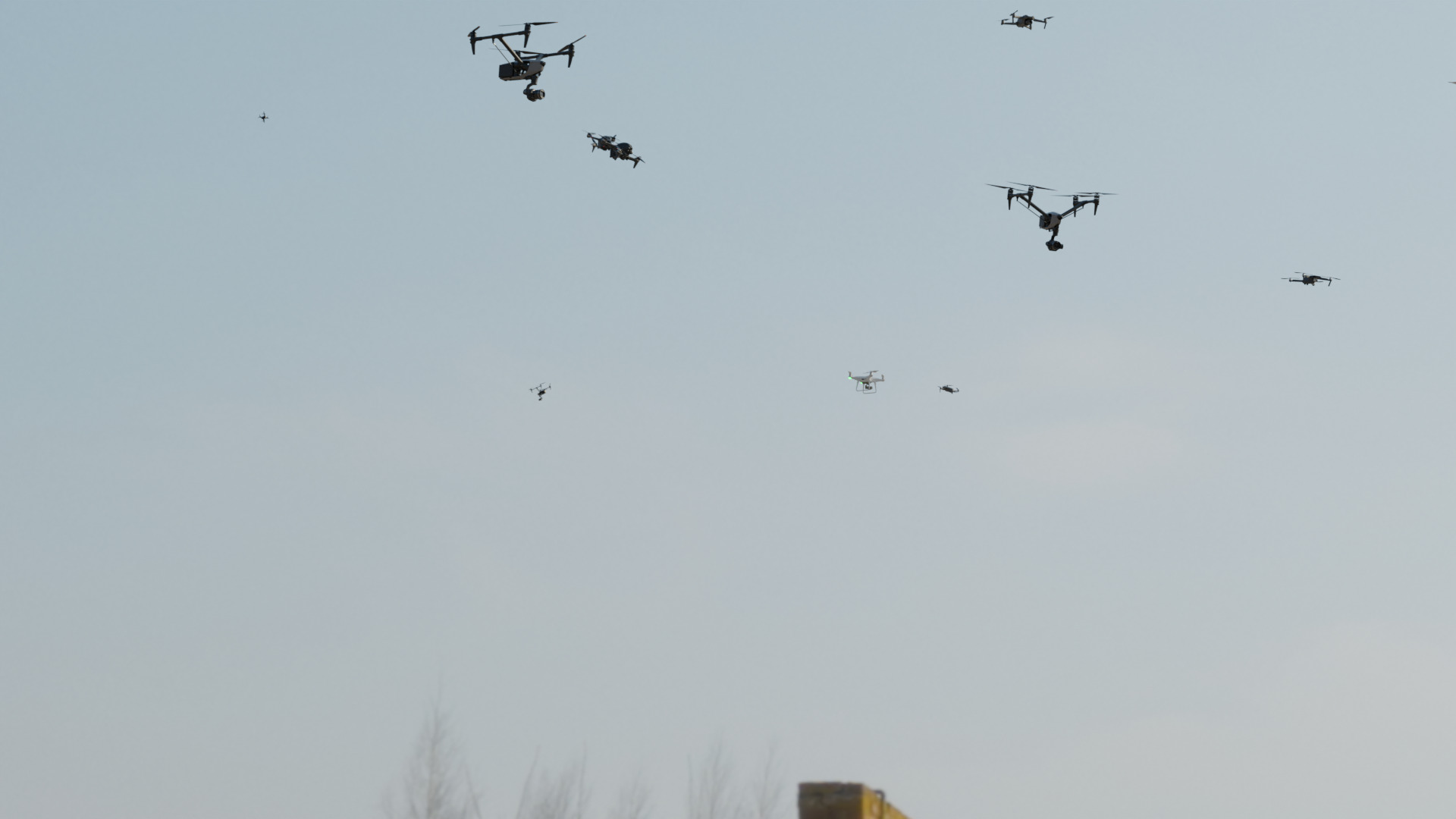}\hfill
        \label{fig:methodology:vanilla_drones}
    }
        \\[\smallskipamount]
    \subfigure[Drones and Birds]{        
        \includegraphics[trim={840px 0 0 0},clip,width=.24\linewidth]{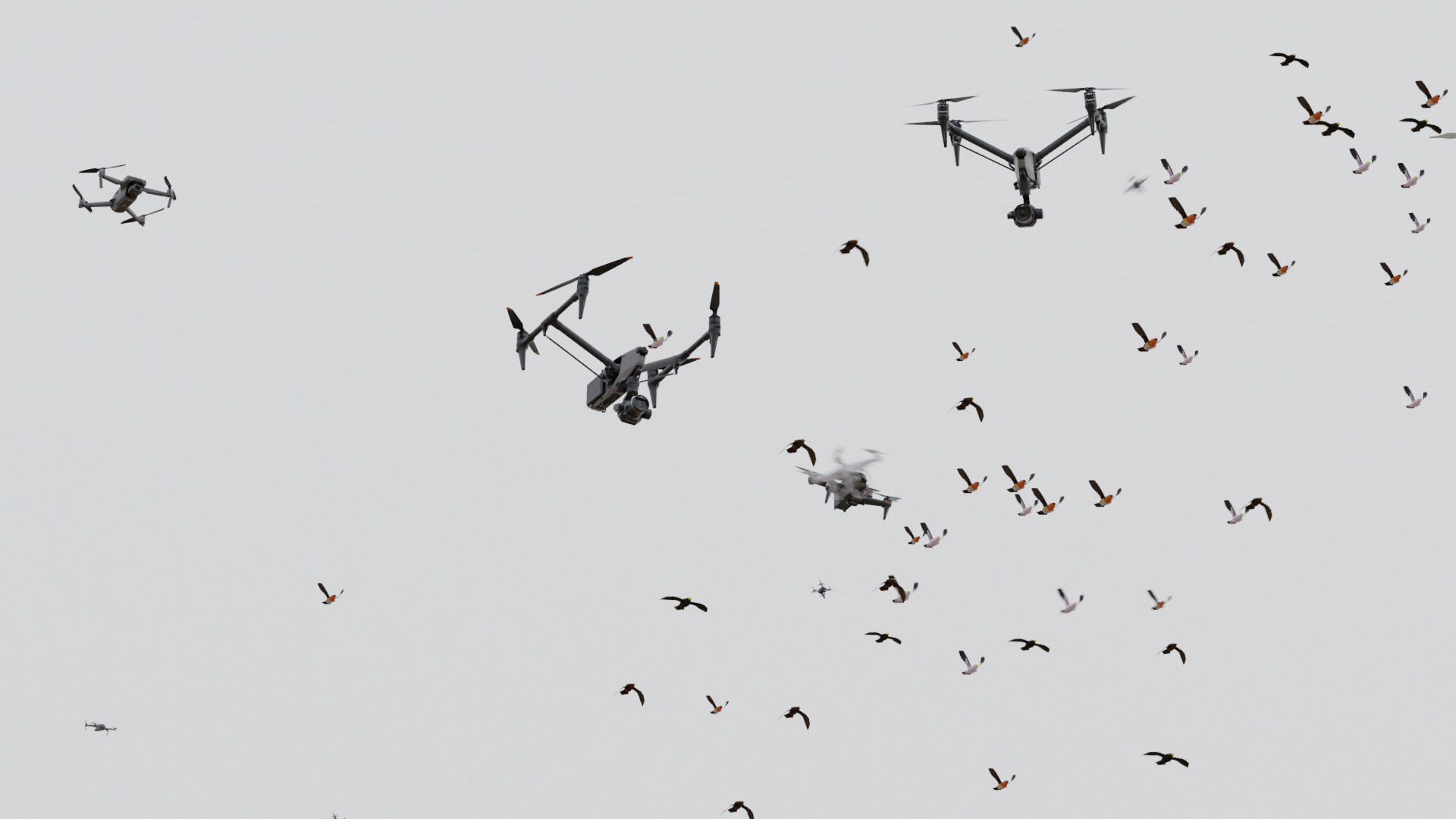}\hfill
        \includegraphics[trim={840px 0 0 0},clip,width=.24\linewidth]{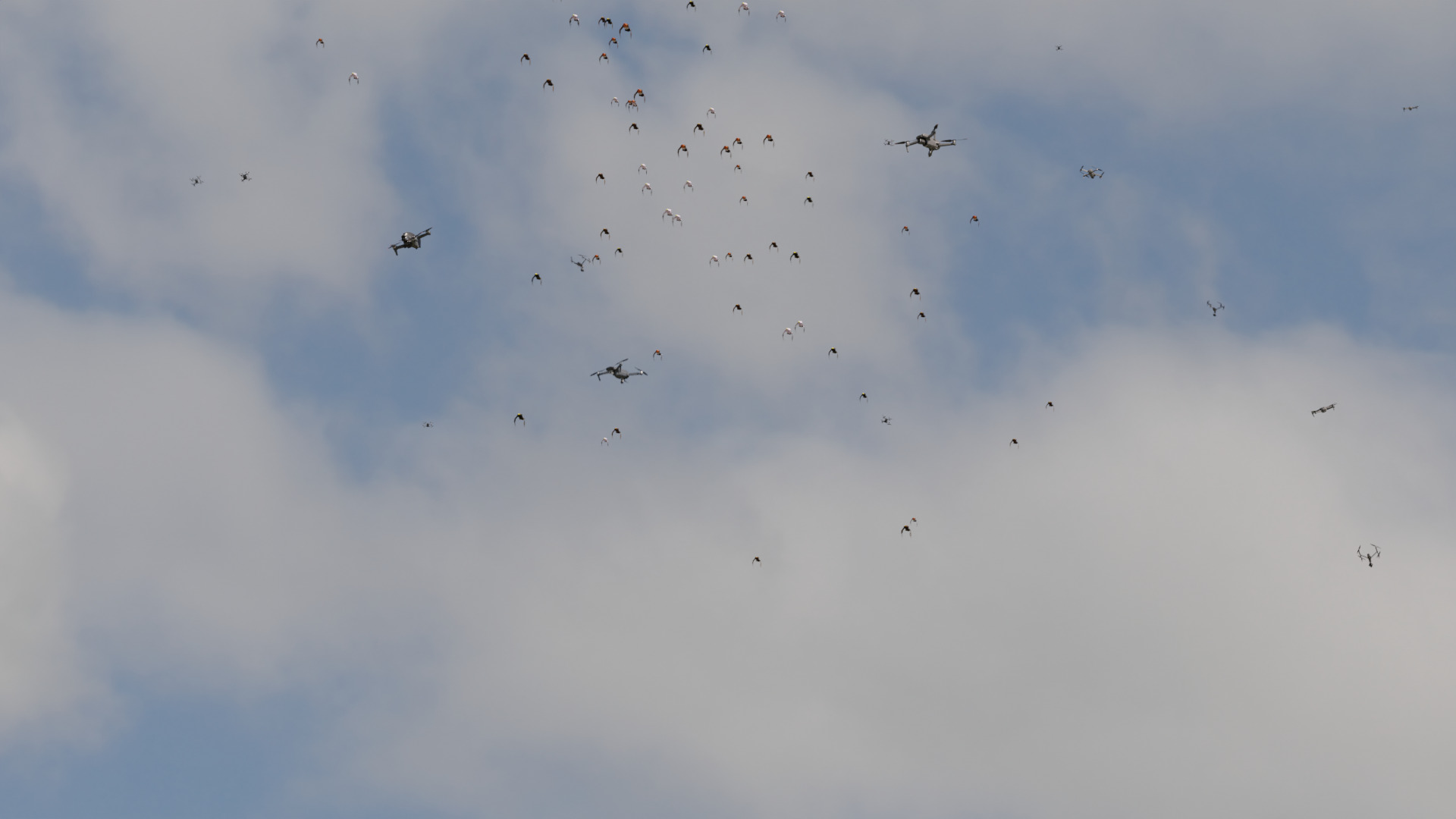}\hfill
        \includegraphics[trim={0 0 840px 0},clip,width=.24\linewidth]{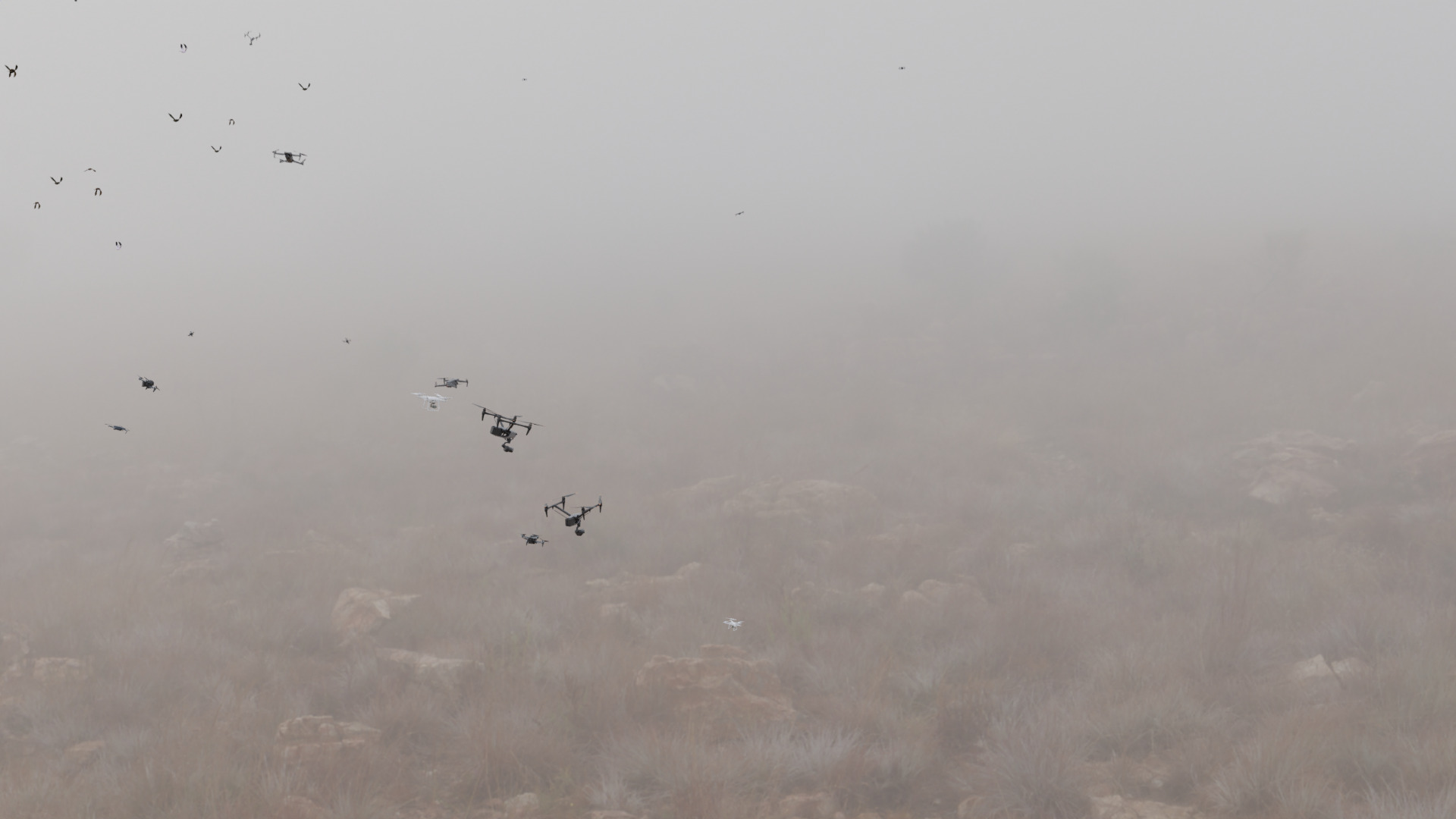}\hfill
        \includegraphics[trim={840px 0 0 0},clip,width=.24\linewidth]{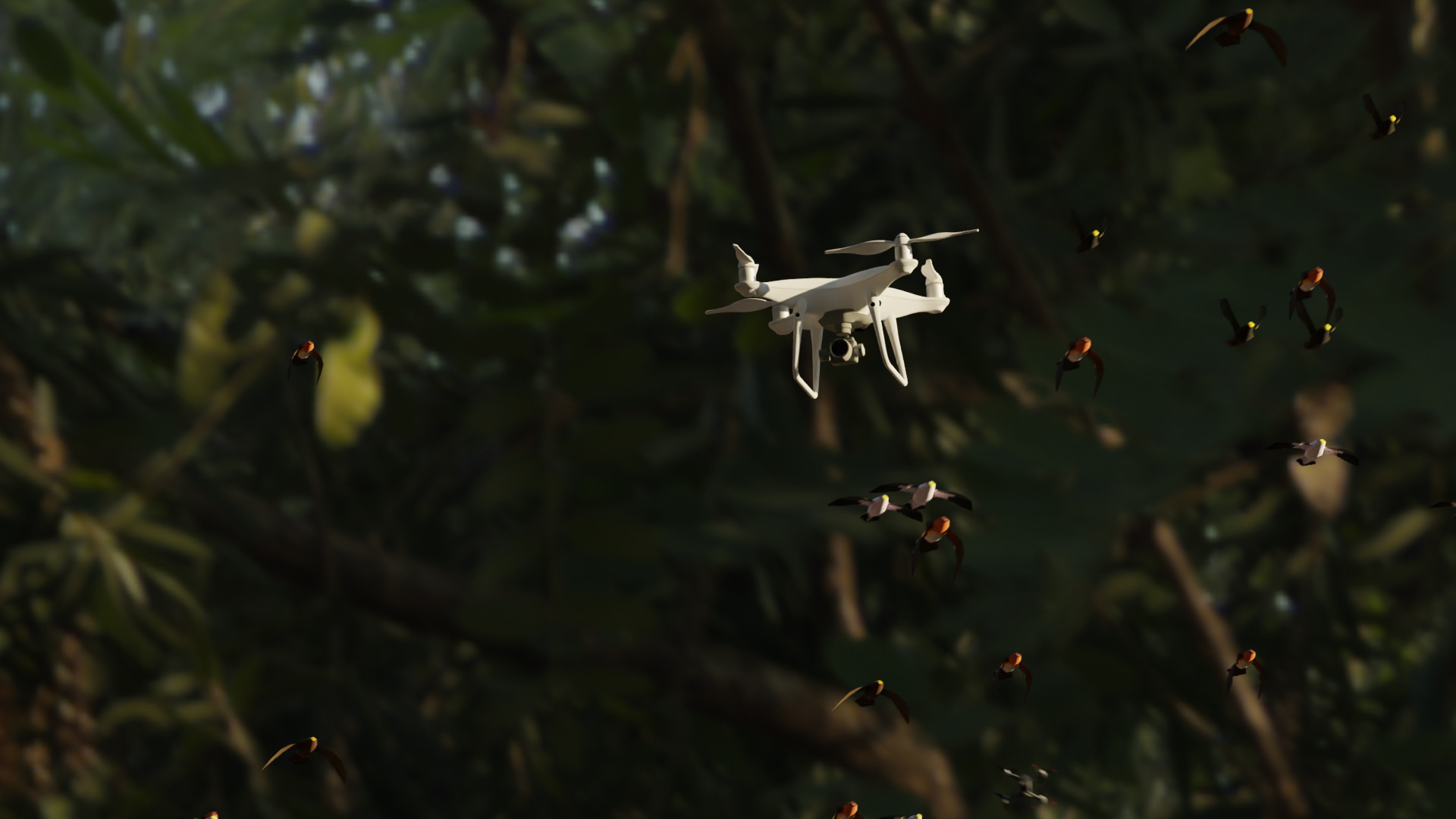}\hfill
        \label{fig:methodology:drones_with_birds}
    }
        \\[\smallskipamount]
    \subfigure[Generic Distractors]{        
        \includegraphics[trim={840px 0 0 0},clip,width=.24\linewidth]{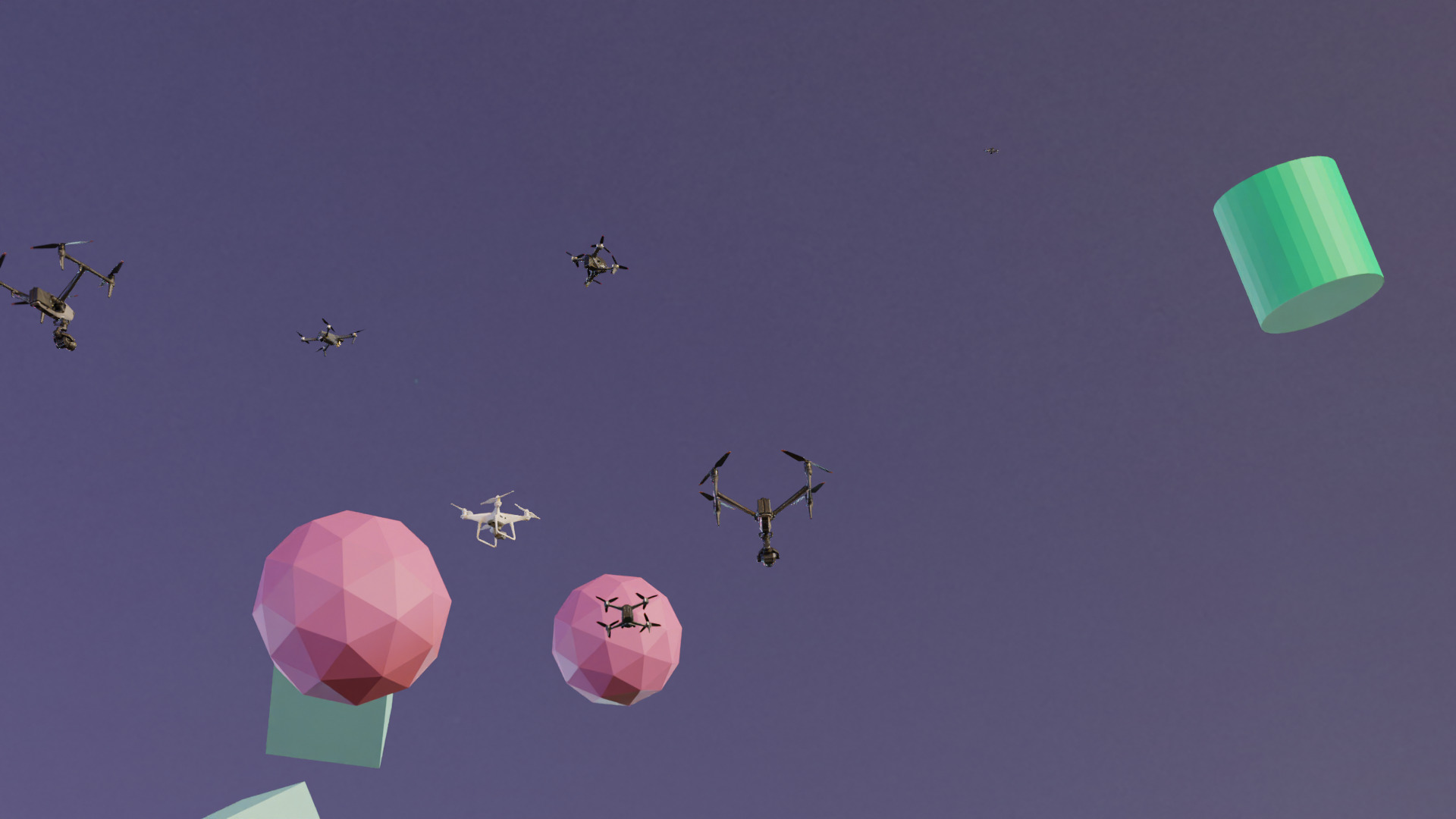}\hfill
        \includegraphics[trim={840px 0 0 0},clip,width=.24\linewidth]{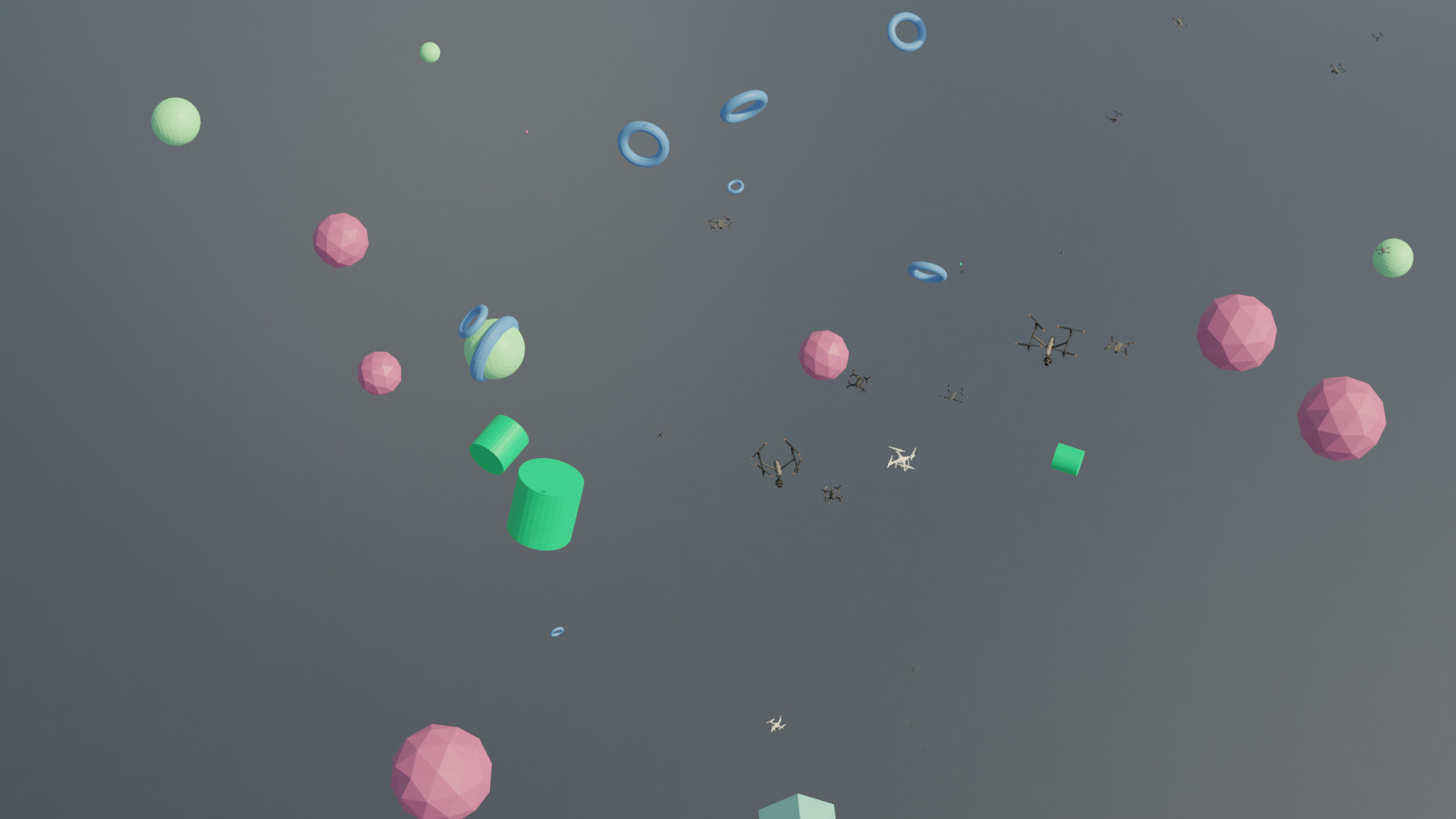}\hfill
        \includegraphics[trim={0 0 840px 0},clip,width=.24\linewidth]{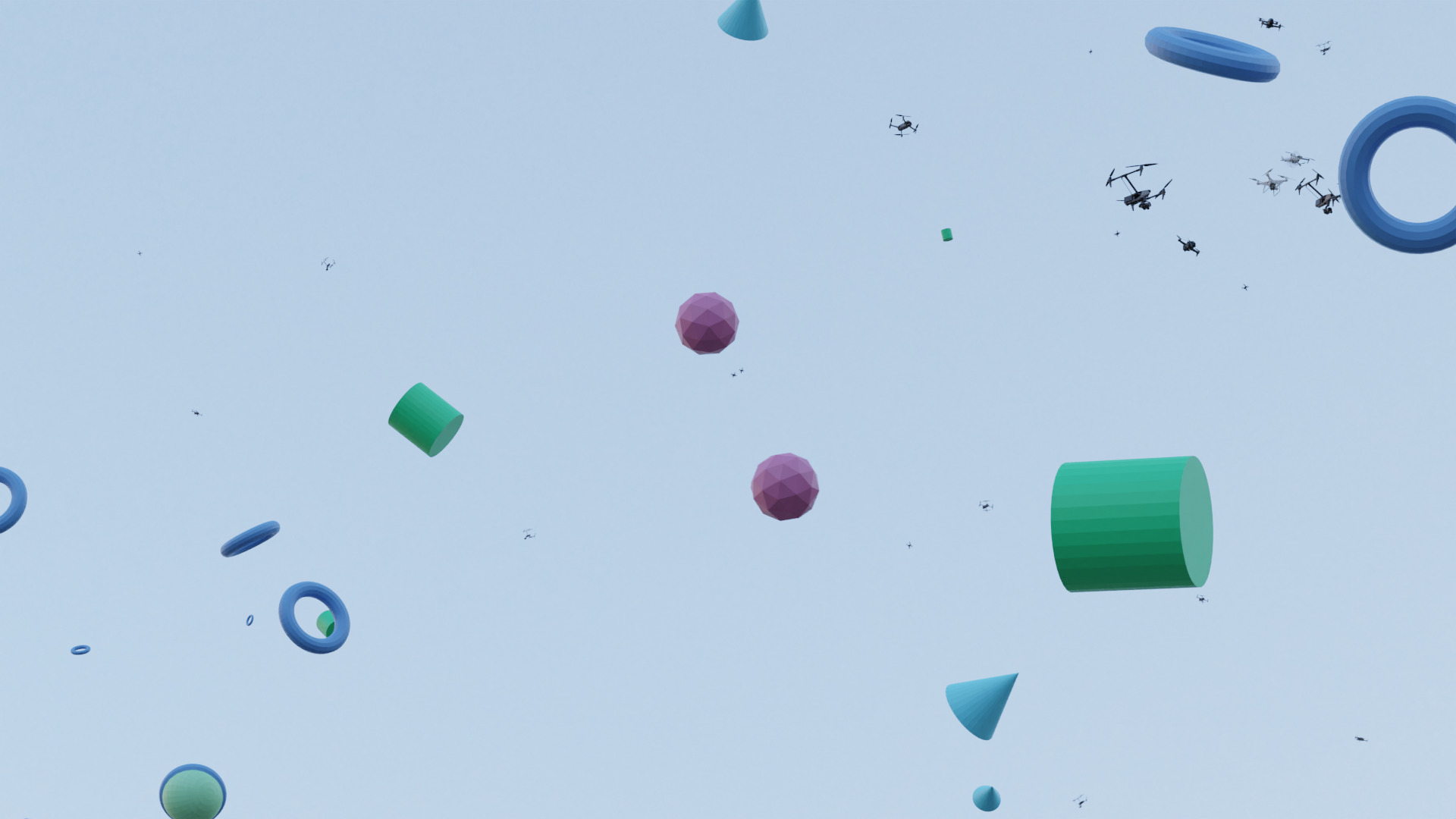}\hfill
        \includegraphics[trim={0 0 840px 0},clip,width=.24\linewidth]{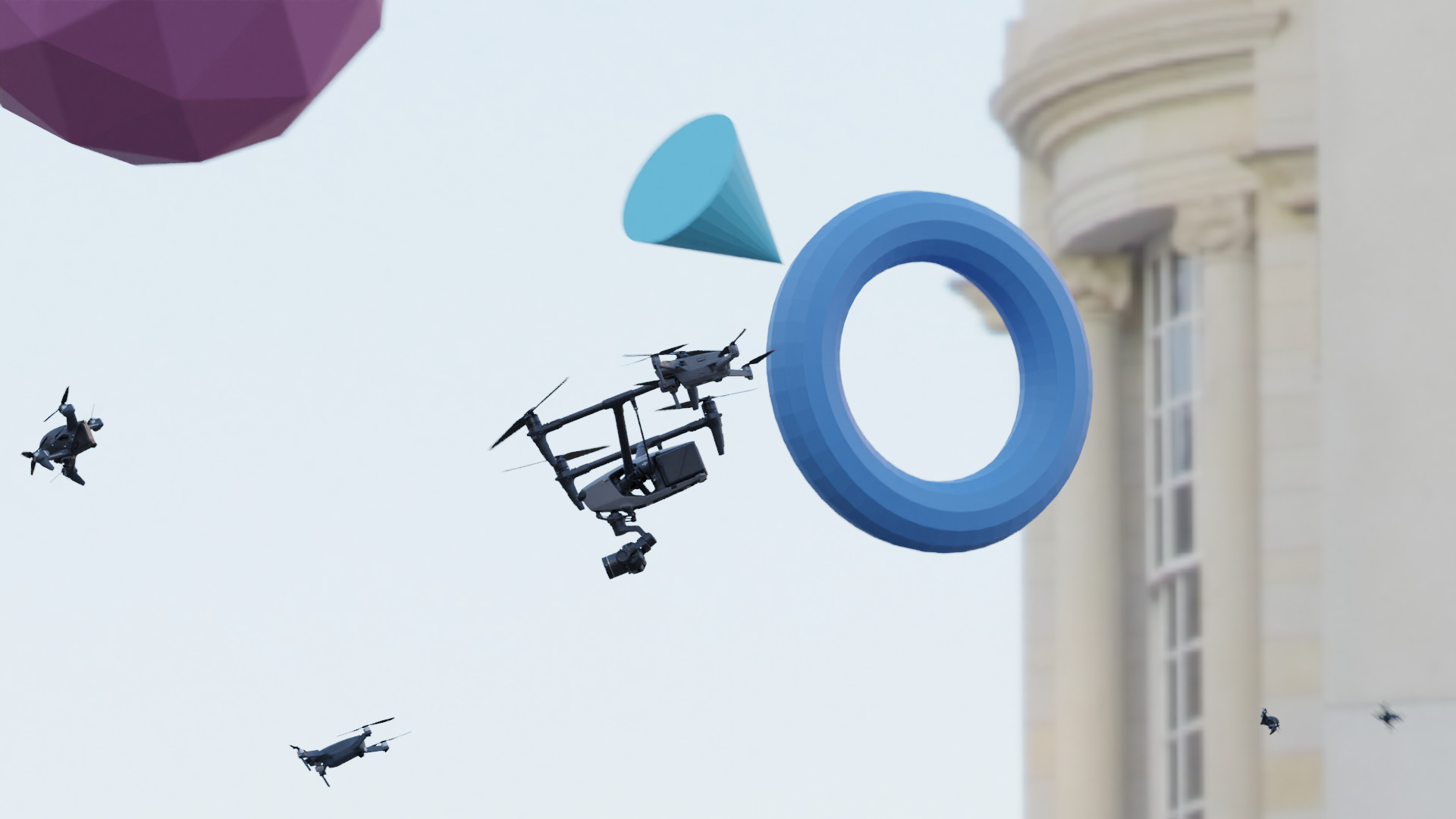}\hfill
        \label{fig:methodology:generic_distractors}
    }
        \\[\smallskipamount]
    \subfigure[Realistic Distractors]{        
        \includegraphics[trim={840px 0 0 0},clip,width=.24\linewidth]{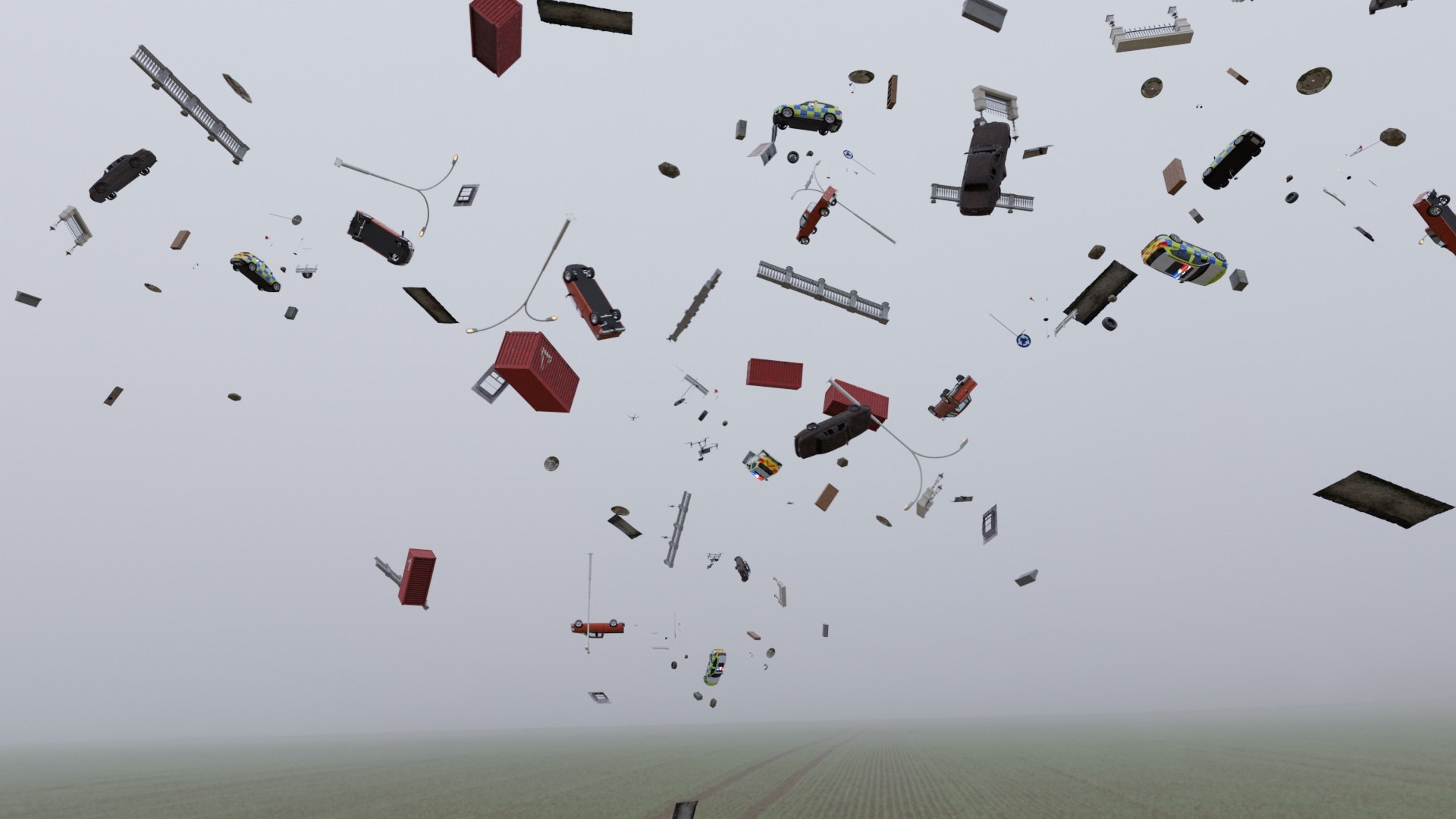}\hfill
        \includegraphics[trim={840px 0 0 0},clip,width=.24\linewidth]{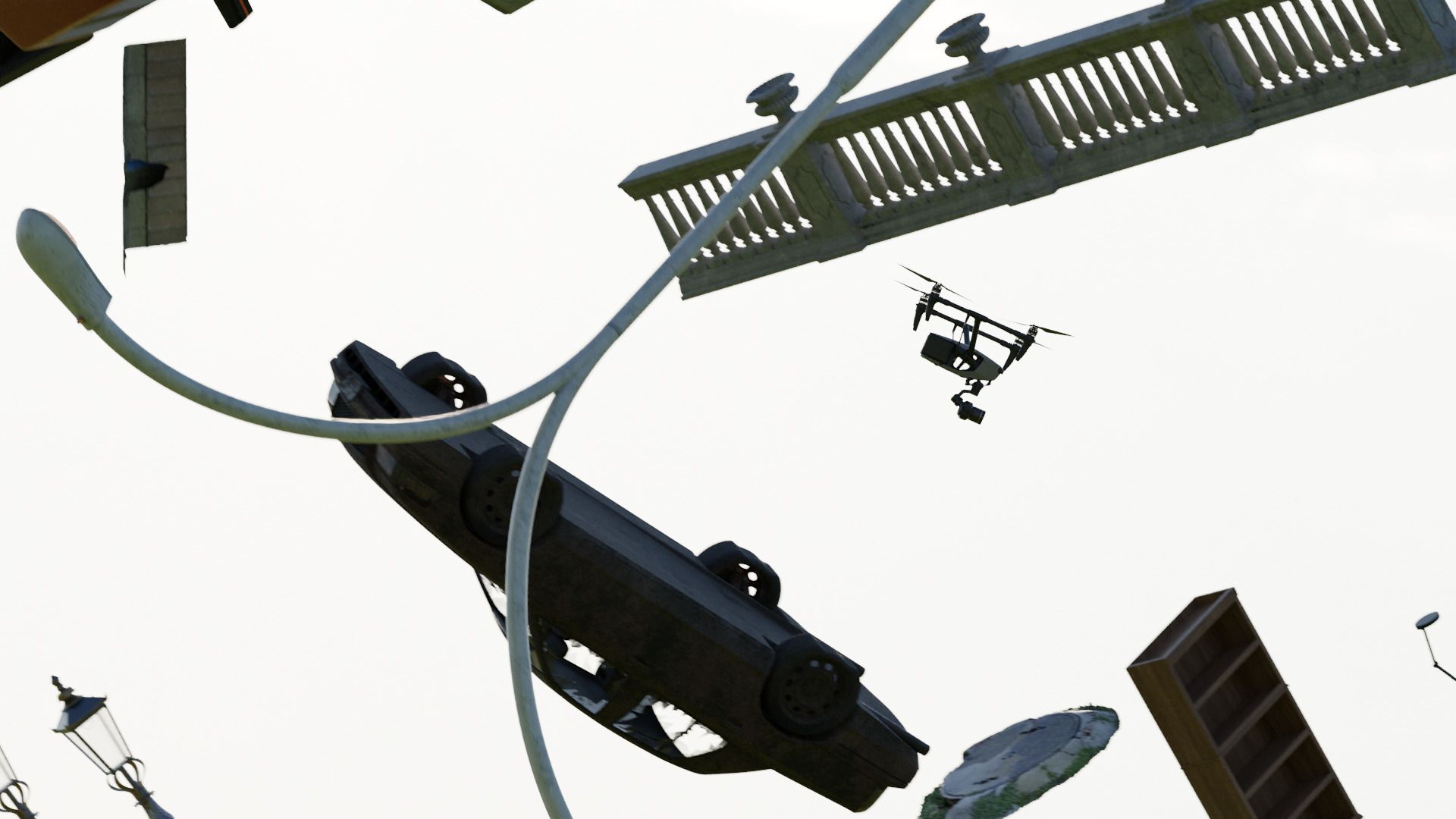}\hfill
        \includegraphics[trim={0 0 840px 0},clip,width=.24\linewidth]{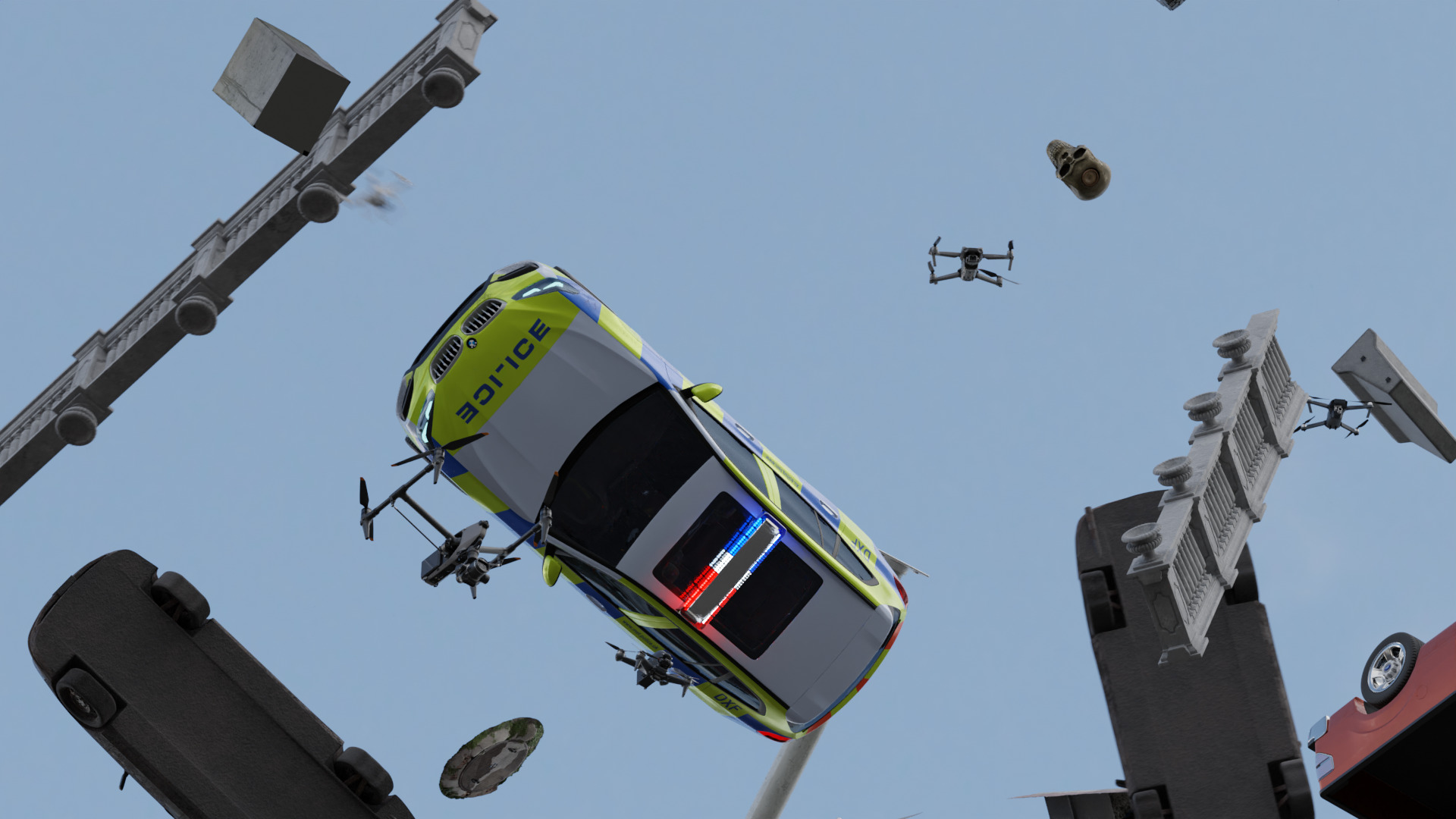}\hfill
        \includegraphics[trim={840px 0 0 0},clip,width=.24\linewidth]{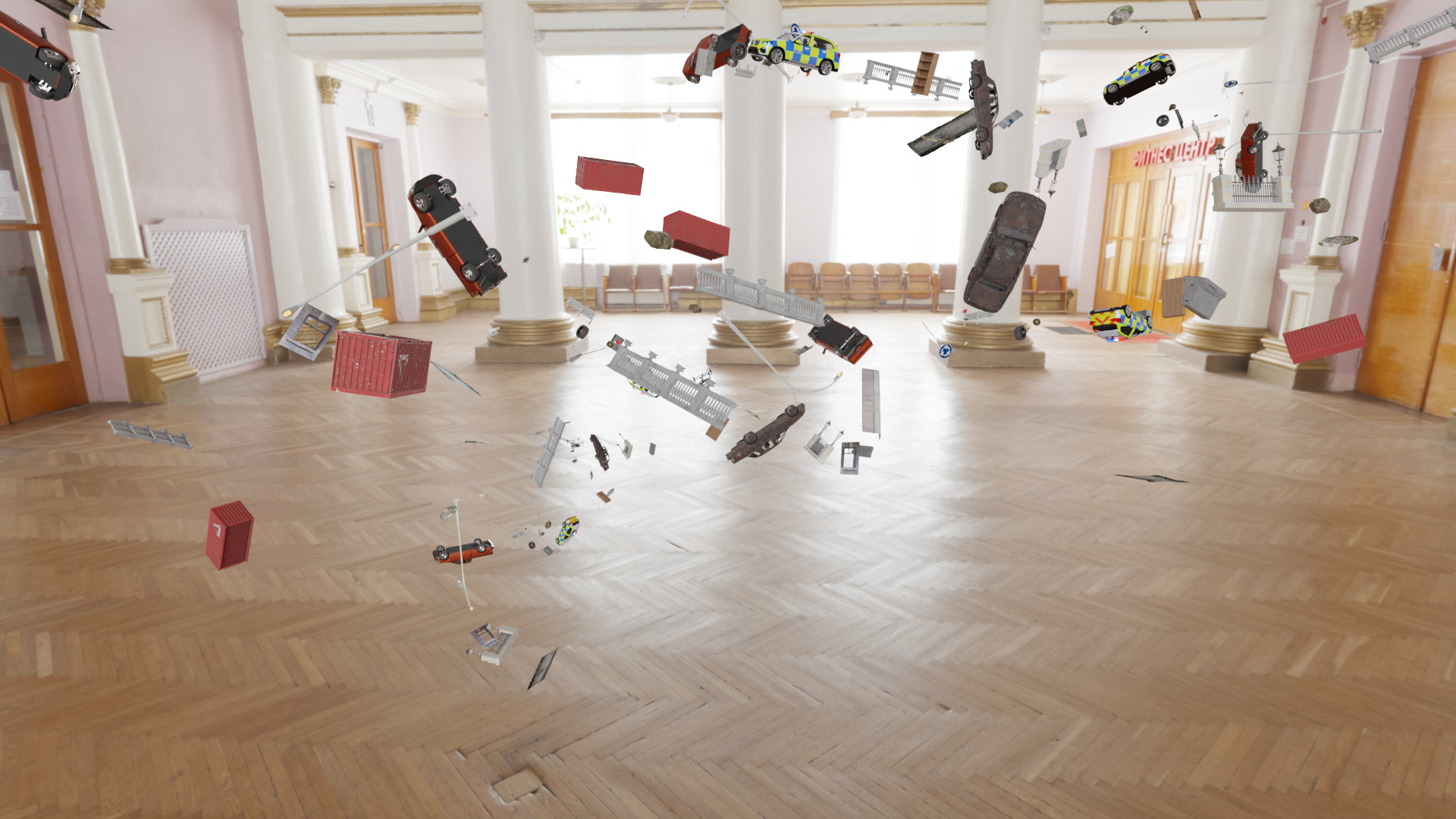}\hfill
        \label{fig:methodology:realistic_distractors}
    }
    \subfigure[Random Backgrounds]{        
        \includegraphics[trim={840px 0 0 0},clip,width=.24\linewidth]{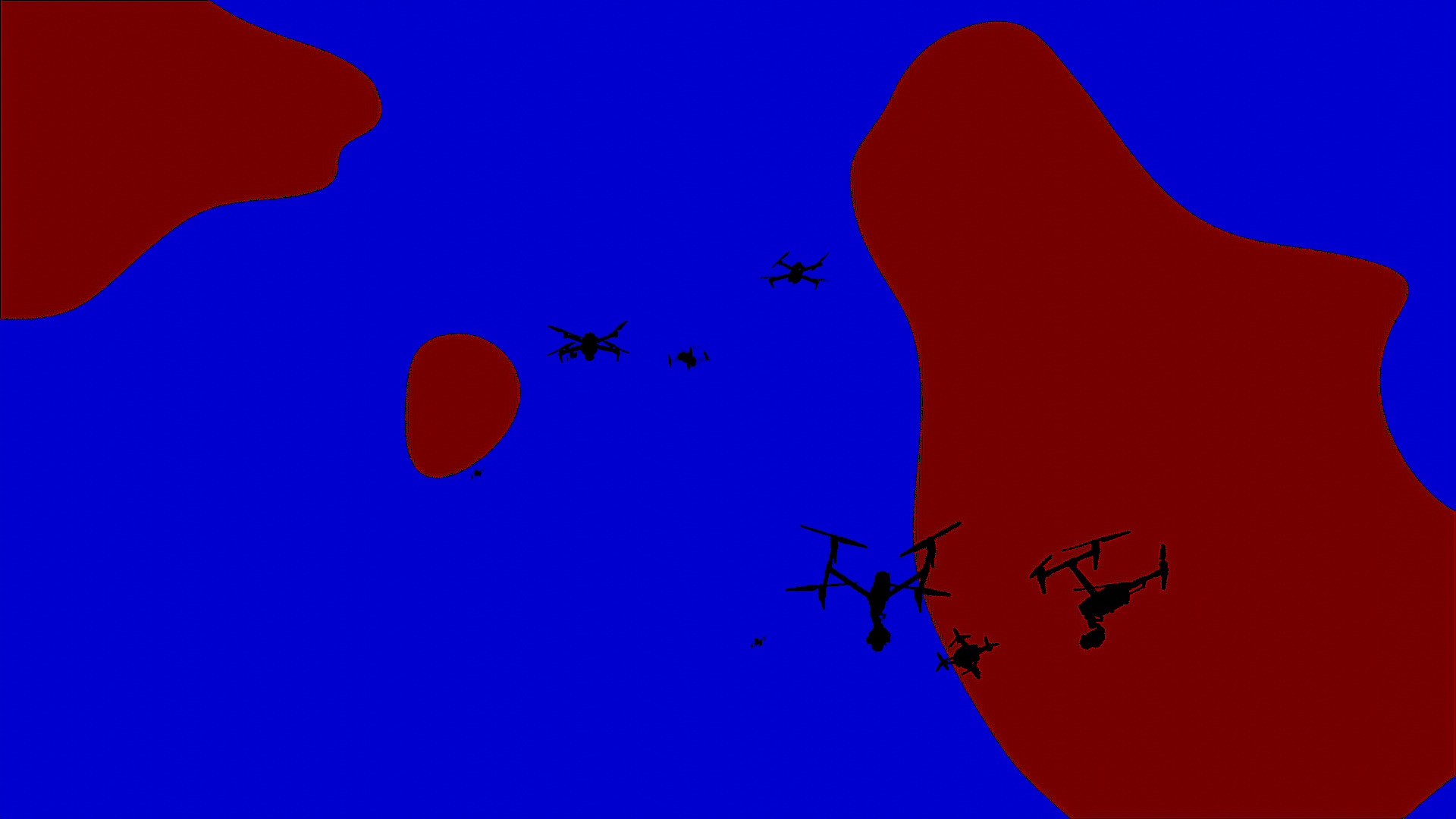}\hfill
        \includegraphics[trim={840px 0 0 0},clip,width=.24\linewidth]{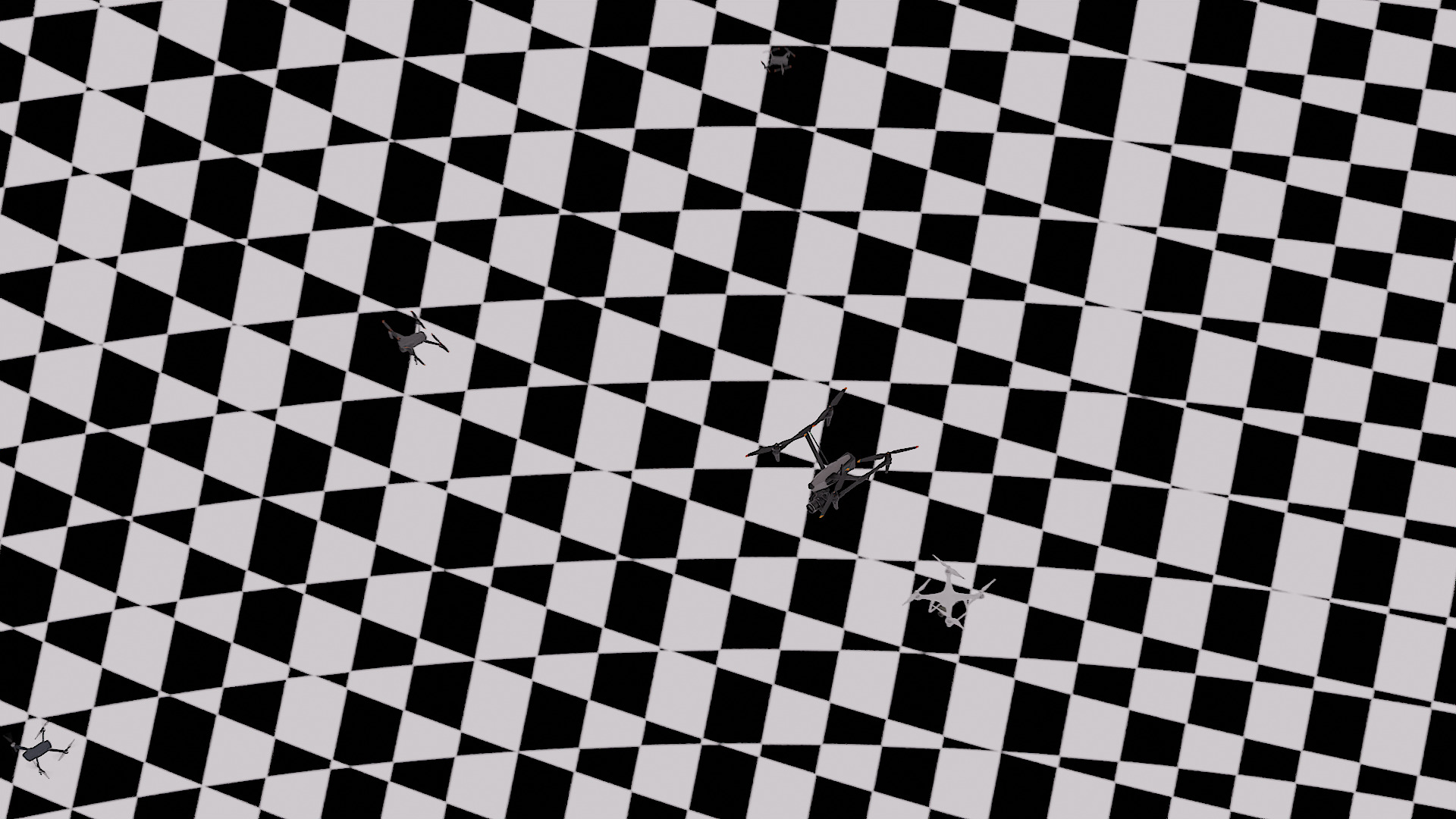}\hfill
        \includegraphics[trim={840px 0 0 0},clip,width=.24\linewidth]{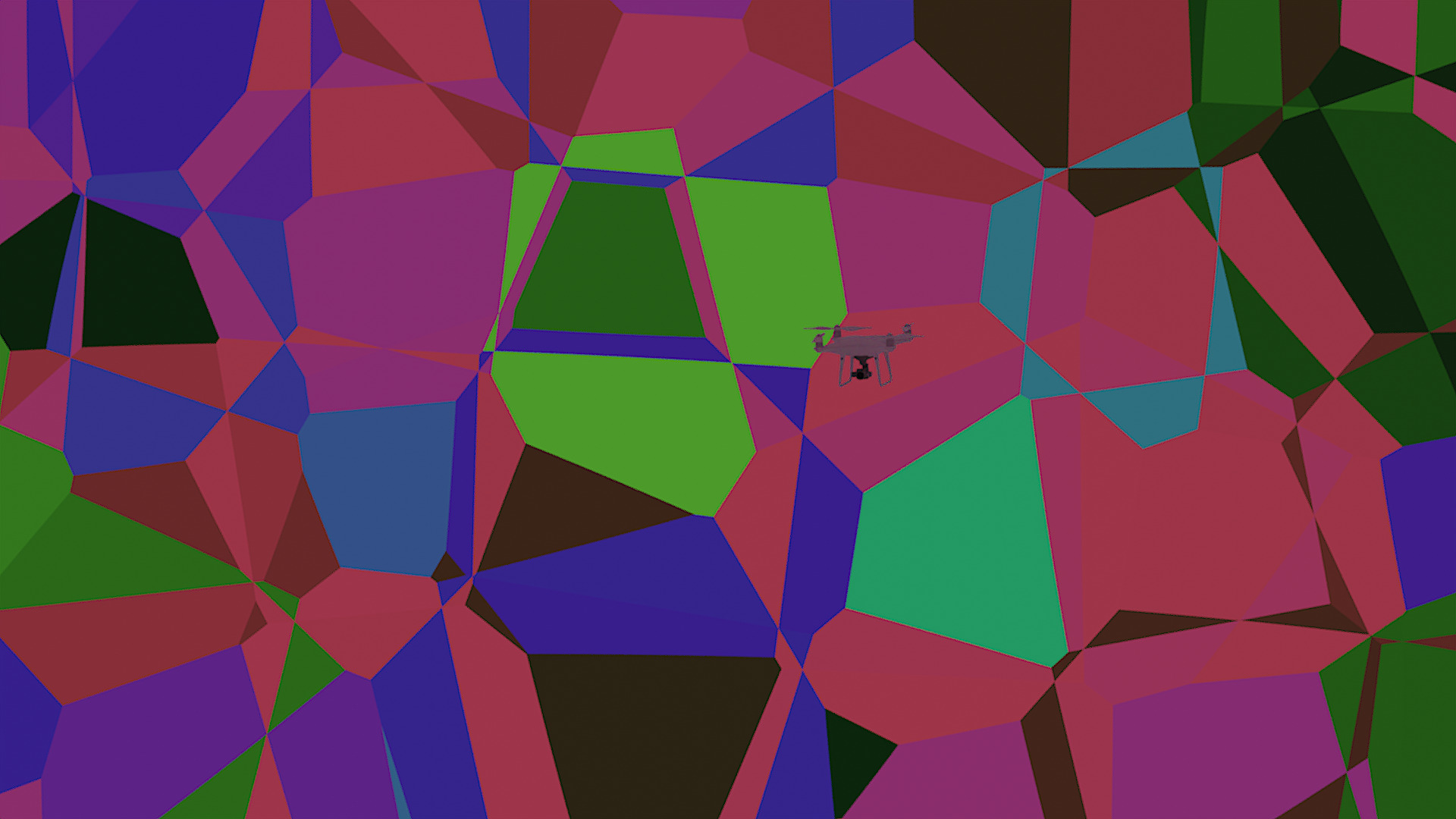}\hfill
        \includegraphics[trim={0 0 840px 0},clip,width=.24\linewidth]{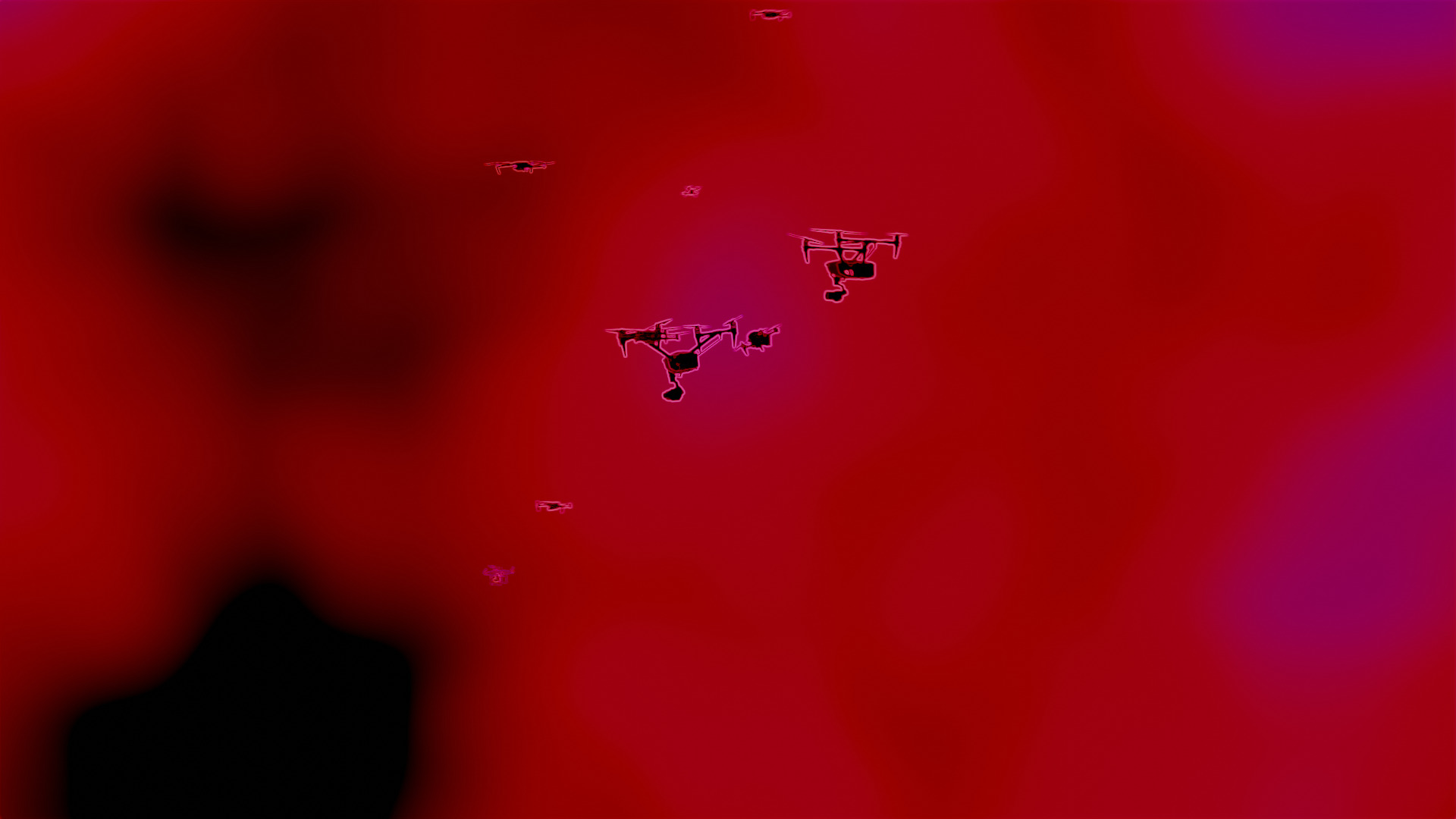}\hfill
        \label{fig:methodology:random_backgrounds}
    }
    \caption{Synthetic datasets generated by using different styles. (\textbf{a}) Drones only, (\textbf{b}) drones and birds, (\textbf{c}) generic distractors, (\textbf{d}) realistic distractors, (\textbf{e}) random backgrounds. }\label{fig:methodology:synth_drones}

\end{figure}

\textbf{Drones Only} (SDR) shown in figure \ref{fig:methodology:vanilla_drones} is a dataset of flying drones. It is the standard version of the dataset as described in section \ref{section:methodology:rendering_process}. 

\textbf{Drones with Birds} (SDR) shown in figure \ref{fig:methodology:drones_with_birds}, is a variation of the drones-only dataset, with the addition of animated birds flying alongside the drones. It is designed as a synthetic version of the Drone-vs-Bird dataset, which contains images of drones with birds acting as distractors. The advantage of the synthetic version is that the birds are labelled in the ground truth mask, unlike in the real dataset, where the birds are not labelled at all. But, for consistency with the Drone-vs-Bird dataset, we will not be using the bird labels during training. 

\textbf{Drones with Generic Distractors} (DR) shown in figure \ref{fig:methodology:generic_distractors}, is a variation of the drones-only dataset, with the addition of generic distractors - cubes, cones, and other basic geometric shapes with random colours added to the scene with the aim of 'distracting' the model. This is based on the findings of Alghonaim \& Johns \cite{alghonaimBenchmarkingDomainRandomisation2021} who report that adding distractors to the synthetic dataset is a key technique for reducing sim-to-real error. This technique is common across domain randomization applications such as Marez et al. \cite{marezUAVDetectionDataset2020} who use distractors for drone detection, and Trembley et al. \cite{tremblayTrainingDeepNetworks2018} who add generic distractors to improve car detection. 

\textbf{Drones with Realistic Distractors} (DR) shown in figure \ref{fig:methodology:realistic_distractors}, is designed as a variation of the generic distractors dataset, with a different style of distractors. The distractors consist of realistic 3D models of objects such as cars, lampposts, cones, traffic signs, etc. (as opposed to geometric shapes in the generic distractors dataset). We label this as a variation of DR and not SDR, because although the objects themselves are realistic, they are not structured in any realistic way.  

\textbf{Drones with Unrealistic Backgrounds} (DR) shown in figure \ref{fig:methodology:random_backgrounds} is a variation of the drones-only dataset, with the realistic HDRI backgrounds replaced by random coloured/patterned backgrounds. This was not done in the style of a particular paper, although Alghonaim \& Johns did experiment with different background types. We hypothesized that if the unrealistic textures and objects can improve the performance of the detector (as shown by the majority of domain randomization literature), random backgrounds should act similarly. The random backgrounds are generated using random colour palettes, with random shapes. Different blending modes are also applied to the images.

\subsection{Neural Network Architecture and Training}\label{section:methodology:neural_network_training}
The primary aim of this study is to compare the results with Isaac-Medina et al. hence any of the architectures they used (Faster R-CNN, SSD, YOLO, DETR) was acceptable. Faster R-CNN was also used in other publications such as Freudenmann et al. \cite{freudenmannExploitationDataAugmentation2021}, hence we chose it as our architecture. Keeping the architecture common allows us to compare the impact of the dataset while controlling for other variables. 

The Faster R-CNN \cite{renFasterRCNNRealTime2016} network is trained using ResNet-50 as a backbone. The network is pretrained on the MS COCO dataset \cite{linMicrosoftCOCOCommon2015}. Unless otherwise specified, the settings of the model are unchanged from the default implementation in PyTorch \cite{paszkePyTorchImperativeStyle, VisionTorchvisionModels}, an SGD optimizer is used with a learning rate of 0.0003, momentum of 0.9, and weight decay of 0.0005. We use a step learning rate scheduler which decays the learning rate by 0.1 every 3 epochs. The networks are trained for 10 epochs. 

\subsubsection{Data Augmentations: Cropping, Noise, and Compression}\label{section:methodology:augmentations}
Augmentations are a common way to improve the diversity of the dataset by performing common operations such as cropping, changing contrast, blurring, etc.
Adding noise to datasets as an augmentation has been tried in synthetic data literature with conflicting reports. 
Some articles report a performance improvement \cite{tremblayTrainingDeepNetworks2018}, while others report it to be negligible \cite{tobinDomainRandomizationTransferring2017}. Wisniewski et al. 
\cite{wisniewskiDroneModelClassification2022} find that adding synthetic noise improves the sim-to-real performance of drone classification. 
The images generated by Blender are in the lossless PNG format. However, the real-life datasets are compressed in the JPEG format, reducing the size, but also reducing the quality. We speculated that this might affect the ability of the network to transfer to real-world datasets. Poyser et al. \cite{poyserImpactLossyImage2020} investigate the impact of compression on the performance of CNNs. They show that, in the task of object detection, a model not trained on compressed imagery drops performance when tested on compressed imagery. However, they find that after retraining the model on compressed imagery the performance improves for even highly compressed images (although the performance of the FRCNN model still drops as the images are compressed more). 

To quantify the effects of this, we perform an experiment where we compare a model trained on only lossless PNG images, with a model where 50\% of the images in the dataset are compressed uniformly in a range of 0-95\%. To achieve this we use the imgaug library \cite{jungImgaug2023}. 

\subsection{Testing: Datasets and Performance Metrics}\label{section:methodology:neural_network_testing}

The goal of this study is to find whether the use of a purely synthetic drone dataset to train the neural network for sim-to-real transfer is feasible. Ideally, the model should be transferrable to many different real-life datasets (and not just a single one). 
This goes against the ethos of neural networks, which assume that the testing dataset should be a subset of the training dataset. 
Instead, we are aiming for out-of-domain generalization, wherein the training datasets and testing datasets originate from different domains, hence bridging the sim-to-real gap. 
To this end, we test our neural network model on three different real-life datasets: MAV-VID, Anti-UAV, and Drone-vs-Bird. We compare our results with a benchmark developed by Isaac-Medina et al., where they trained a model for each of the datasets, and tested each model on a subset of the dataset that they were trained on. Instead, we propose a single set of weights which is tested on unseen real-life datasets.

We aim for the sim-to-real to be accurate, and we also aim for it to be generalizable - we want the weights to perform well on any drone dataset. 
We use the publicly available real-world datasets used by Isaac-Medina et al. in their benchmark: MAV-Vid, Drone-vs-Bird, and Anti-UAV.

\textbf{MAV-VID} \cite{rodriguez-ramosAdaptiveInattentionalFramework2020} consists of 29,500 images used for training and 10,732 images for validation. It contains videos of drones captured by other flying drones and drones captured from a ground camera.

\textbf{Drone-vs-Bird} \cite{colucciaDronevsBirdDetectionChallenge, colucciaDroneVsBird2021} consists of 85,904 images used for training and 28,856 images for validation. The videos are taken from static cameras near backdrops of buildings, fields, lakes, and landscapes. The aim of the dataset is to reduce false positives with regard to birds coming up as positive objects instead of drones. However, the birds are not labelled in the dataset (only drones are).

\textbf{Anti-UAV} \cite{jiangAntiUAVLargeMultiModal2021} consists of 149,478 images used for training and 37,016 images for validation. It features videos of drones captured by a pan-tilt-zoom (PTZ) security camera, on a backdrop of cities. The PTZ camera is manually operated to track the drones. The dataset features videos from an optical camera in the visible spectrum and from an infrared (IR) camera. We will focus on the visible spectrum images only. Although its aim is to benchmark UAV tracking capabilities, we can still use it to benchmark drone detection.

The datasets described contain only drone labels and do not contain labels for other objects (such as birds, or other flying objects). COCO metrics (that we will be using for performance metrics in this chapter) work by finding the Intersection over Union between the prediction and the ground truth. 

\begin{figure}
\centering
\includegraphics[width=0.3\textwidth]{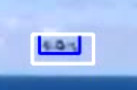}
\caption{A zoomed-in example of an inaccurate ground truth on the MAV-Vid dataset. Although the model prediction (blue rectangle) is more accurate than the ground truth (white rectangle), it falls below the intersection over union (IoU) of 0.5 required for a correct prediction. \label{fig:methodology:bad_ground_truth}}
\end{figure}  

The accuracy of the manual ground truth labels within the datasets varies in quality. Figure \ref{fig:methodology:bad_ground_truth} shows an example of a ground truth (white rectangle) which is inaccurate. In this case, the model prediction (blue rectangle) is more accurate than the ground truth, but because it falls below the intersection over union (IoU) of 0.5 it is not counted as a correct prediction. This is generally the case, especially for smaller instances of drones. Because it is hard to quantify how many of the ground truth labels are inaccurate, and costly to remedy, we will take the labels as they are. 

\subsection{Repeatability}\label{section:methodology:repeatability}
The process of NN training inherently contains randomness, hence repeating training with the same settings produces different results. This is rarely presented in results, with most publications that we have observed simply presenting the results from a single run. The initial seed has an effect on the results of the training, as presented by Picard \cite{picardTorchManual_seed34072023}. Picard shows that the distributions generated from random seeds generally follow a normal distribution (even though some of their results do appear skewed). Although he mentions \textit{black swan} events \cite{taleb2007black}, we believe they meant this in the context of extreme values on the sides of the Gaussian distribution as opposed to a fat-tailed distribution. Hence, we assume the results from our training runs will follow a Gaussian distribution. To report our results we will take 8 samples at each configuration and report the mean alongside the 95 percent confidence interval.

\section{Results}\label{section:Results}
This section shows the results of the Faster-RCNN object detection architecture trained on a purely synthetic dataset of drones generated in Blender described in section \ref{section:methodology:synthetic_dataset_generation}. The trained model is tested on three separate datasets: MAV-Vid, Drone-vs-Bird (DvB), and Anti-UAV described in section \ref{section:methodology:neural_network_testing}. The aim of these experiments is to understand whether sim-to-real, the process of transferring a model trained on synthetic data to real-world scenarios, is possible for drone detection, and if so, which variables impact the sim-to-real transfer.

We perform the following studies: camera bounds randomization (section \ref{section:results:Bounds_Study}), dataset augmentations (section \ref{section:results:data_augmentations}, dataset size (section \ref{section:results:dataset_size}), and domain randomization styles (section \ref{section:results:domain_randomization_styles}). Lastly, we compare our results to literature in section \ref{section:results:comparion_to_literature}.
Unless otherwise specified, the experiments use the training parameters described in section \ref{section:methodology:neural_network_training}, a camera bounding size of 40 m is used, a dataset size of 5,000 is used, JPEG compression is enabled, noise is enabled, and each experiment is repeated 8 times to produce a mean and a 95 per cent confidence interval. 

\subsection{Camera Bounds Randomization}\label{section:results:Bounds_Study}
The position of the camera with respect to drones is one of the parameters randomized in the simulation. The size of camera bounds in turn is correlated to the size of the drones in the images in the dataset. Smaller camera bounds translate into a higher proportion of large-sized drones, and larger camera bounds translate into a higher proportion of small-sized drones in the dataset. Smaller images of drones may make it harder for the network to learn their shape, but may also improve the performance of the network on datasets which contain drones further away from the camera. Hence it is important to find a balance between the two. To understand the relationship between the bounds of the camera position randomization and the performance on the real-world data, we create 5 datasets with different bounds: 20 m, 40 m, 80 m, 160 m, and 320 m.

\begin{figure}[H]
\centering
\includegraphics[width=0.75\linewidth]{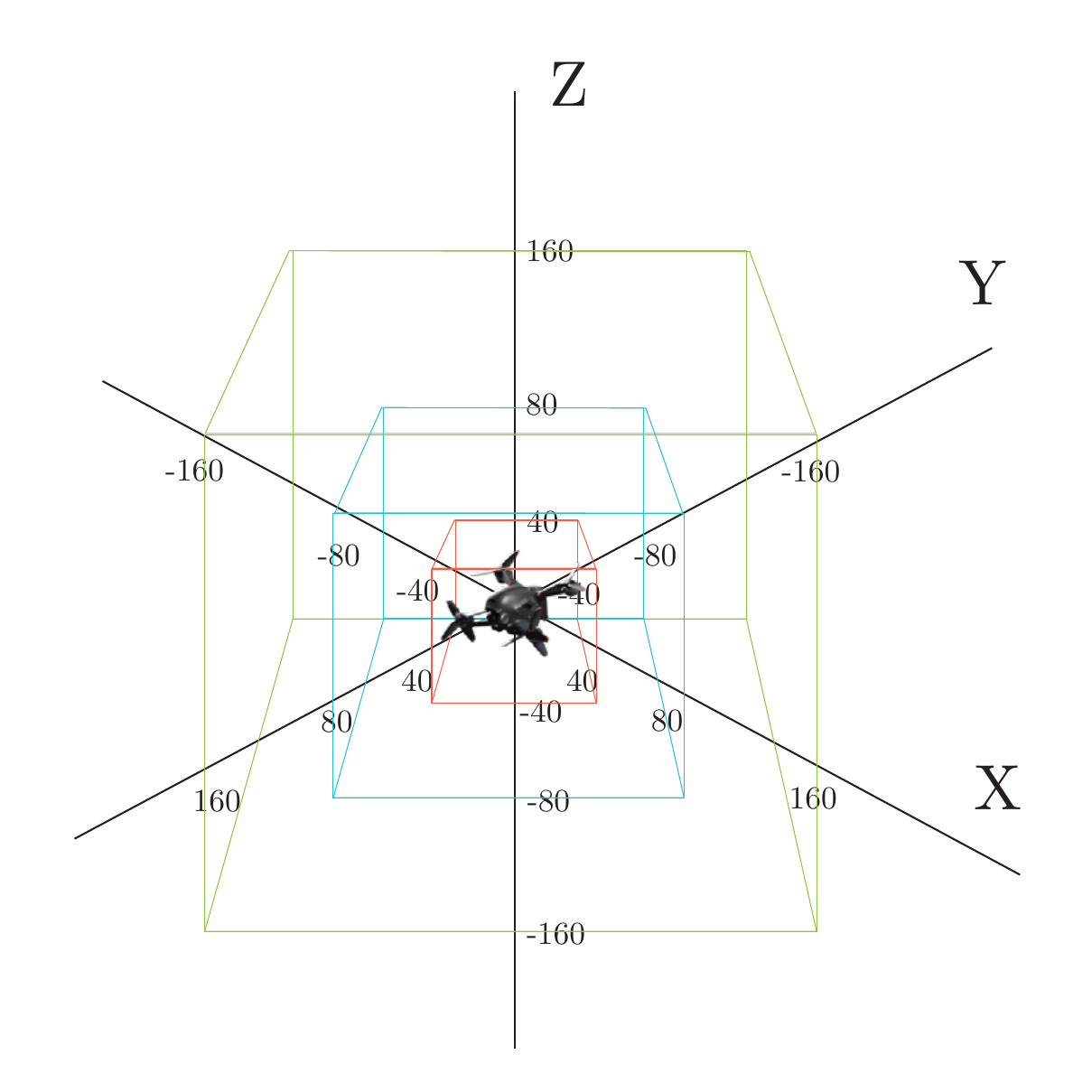}
\caption{A representation selected bounds within which the camera position is randomized. Red: 40 m, blue: 80 m, green: 160 m. Each frame, the position of the camera is randomized within these bounds. We also present results from 20 m and 320 m bounds, which are not illustrated here.  \label{fig:results:camera_bounds}}
\end{figure}

Figure \ref{fig:results:camera_bounds} shows the bounds within which the camera position is randomized. Some of the bounds, red: 40 m, blue: 80 m, and green: 160 m are shown. The bounds of 20 m and 320 m are not shown in the illustration but are used in the comparison results. Each of the bound sizes is used to create a distinct dataset to train a NN model. The focal length is randomized between 15 mm and 300 mm, and this range stays fixed throughout. All of the other parameters stay fixed for this experiment. An ablation study is performed by training the NN model on each of the datasets. 

\begin{figure}
\centering
\includegraphics[width=1.0\linewidth]{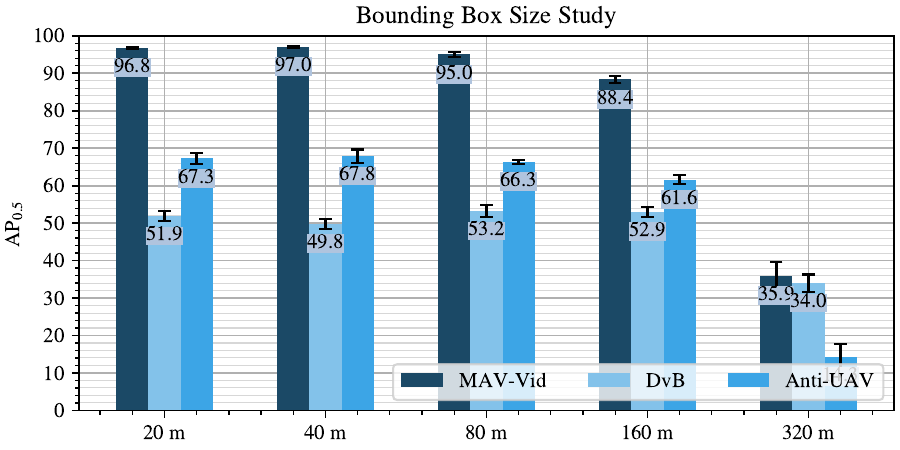}
\caption{Results of the bounding size study for 20 m, 40 m, 80 m, 160 m, and 320 m bounds. Average Precision at .50 IoU ($AP_{0.5}$) on the y-axis, presented as a mean of 8 runs along with a 95 percent confidence interval, and the selected bounding size on the x-axis. Each bounding size is tested on the MAV-Vid, Drone-vs-Bird (DvB), and Anti-UAV datasets.  \label{fig:results:bounding_size_results}}
\end{figure}  

Figure \ref{fig:results:bounding_size_results} shows the results of the bounding size ablation study. 20 m, 40 m, and 80 m bounding sizes all show reasonably similar results across the three different test datasets, with a mean $AP_{0.5}$ ranging from 95.0\% to 97.0\% on the MAV-Vid dataset, 49.8\% to 53.2\% on DvB, and 66.3\% to 67.8\% on Anti-UAV. 160 m shows a significant drop on the MAV-Vid dataset down to a mean of 88.0\% $AP_{0.5}$, but remains competitive on the DvB dataset, showing a mean $AP_{0.5}$ of 53.6\%. This is likely because the DvB dataset contains more images of smaller drones and larger camera bounds produce more images of smaller drones. Hence, it seems that the model benefits from a higher proportion of smaller drones for higher DvB performance, but suffers on MAV-Vid as a result. At 320 m, the model struggles to learn the shape of the drones the performance significantly drops across all 3 datasets. This study has shown that the bounds do not have a strong effect on the sim-to-real transferability if kept within reasonable bounds between 20 m and 80 m. The peak performance also varies across the 3 real-world datasets. Both MAV-Vid and Anti-UAV perform best on the 40 m dataset, while the DvB performance peaks between 80 m and 160 m. 

\subsection{Data Augmentations (JPEG Compression and Noise)}\label{section:results:data_augmentations}
We investigate the effects of JPEG compression and Gaussian noise data augmentations, as described in \ref{section:methodology:augmentations}. 

\begin{figure}
\centering
\includegraphics[width=1.0\linewidth]{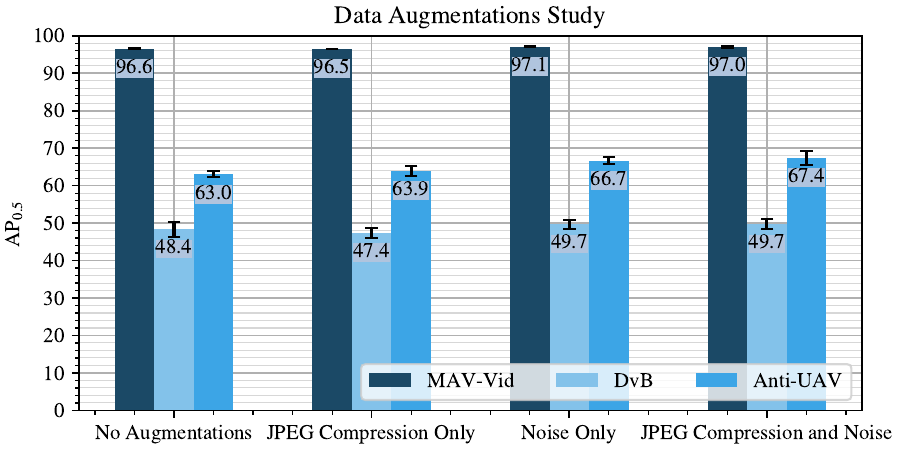}
\caption{Results of the data augmentations study. Average Precision at .50 IoU ($AP_{0.5}$) on the y axis, presented as a mean of 8 runs along with 95 percent confidence interval, and the data augmentation on the x axis. Each model is tested on the MAV-Vid, Drone-vs-Bird (DvB), and Anti-UAV datasets.  \label{fig:results:data_augmentations}}
\end{figure} 

Figure \ref{fig:results:data_augmentations} shows the results of the data augmentations study, testing whether JPEG compression and noise have any effects on the sim-to-real transferability. The dataset with no augmentations applied achieves a mean $AP_{0.5}$ of 96.6\% on MAV-Vid, 48.4\% on DvB, and 63.0\% on the Anti-UAV dataset. 
When applying JPEG compression, the results do not show a significant difference. Adding noise seems to have a slight effect, producing a mean $AP_{0.5}$ of 97.1\% on MAV-Vid, 49.7\% on DvB, and 66.7\% on the Anti-UAV dataset. Adding both JPEG compression and noise does not produce a significant difference compared with just adding noise. This suggests that JPEG compression has no noticeable effect on sim-to-real transferability. Adding noise has a weak effect, slightly improving the mean $AP_{0.5}$ across the three datasets. 

\subsection{Dataset Size}\label{section:results:dataset_size}
We investigate the effects of the dataset size on the performance of the model. We consider 8 different sizes: 50, 100, 250, 500, 1,000, 2,500, 5,000, and 15,000 images. The 15,000 image dataset is concated using the 20 m, 40 m, and 80 m datasets as described in section \ref{section:results:Bounds_Study}.

Unlike in other tests, we did not perform a repeat of these experiments (apart from 5,000 and 15,000 images). Further, a limitation of this study is the 15,000 dataset as it is compiled from the different bounds datasets. However, because each of the standalone datasets presented similar results, by combining them we should notice size differences, if any. 

\begin{figure}
\centering
\includegraphics[width=1.0\linewidth]{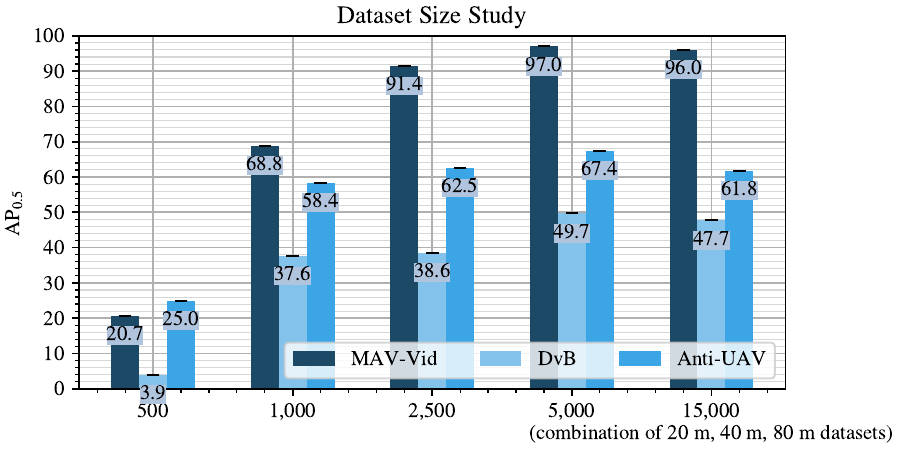}
\caption{Results of the dataset size study for 50, 100, 250, 500, 1,000, 2,500, 5,000, and 15,000* images. Average Precision at .50 IoU ($AP_{0.5}$) on the y-axis, and the selected dataset size on the x-axis. Each bounding size is tested on the MAV-Vid, Drone-vs-Bird, and Anti-UAV datasets. *Note that the 15,000 is composed by concatenating the 20 m, 40 m, and 80 m datasets from the bounds study.   \label{fig:results:dataset_size_results}}
\end{figure}  

Figure \ref{fig:results:dataset_size_results} shows the results of the dataset size study. As expected, larger datasets result in better results. This appears to taper off at around 5,000 images, with no significant difference found between a dataset size of 5,000 and 15,000. A limitation of this study is that a single learning rate is used. Larger datasets may benefit from different learning rates.
This result is consistent with the studies performed by Trembley et al. \cite{tremblayTrainingDeepNetworks2018}, who find that above a training dataset size of 10,000, the results remain consistent (roughly +/- 1\% AP at .50 IoU for a pre-trained model between a dataset size of 10,000 and 900,000). However, their study is for a different application of car detection with different training and test datasets, with different training parameters, although they do use domain randomization for their training dataset. Hence, it is possible that the similarity is a coincidence.

\subsection{Domain Randomization Styles}\label{section:results:domain_randomization_styles}

Domain randomization styles can be implemented in many different ways, with no generally accepted consensus, and a lack of detailed quantitative studies. The problem is made harder because the findings are not necessarily transferable to different applications. As discussed in the literature section, although some studies utilize domain randomization for drone detection, we found a lack of quantitative data comparing what works and what does not. In section \ref{section:methodology:variation_in_datasets} we describe the process of generating datasets using different styles of domain randomization. We presented the following datasets: drones only, drones with birds, drones with generic distractors, drones with realistic distractors, and drones with unrealistic backgrounds. We train the Faster-RCNN model using each of the datasets and compare the results. 

\begin{figure}
\centering
\includegraphics[width=1.0\linewidth]{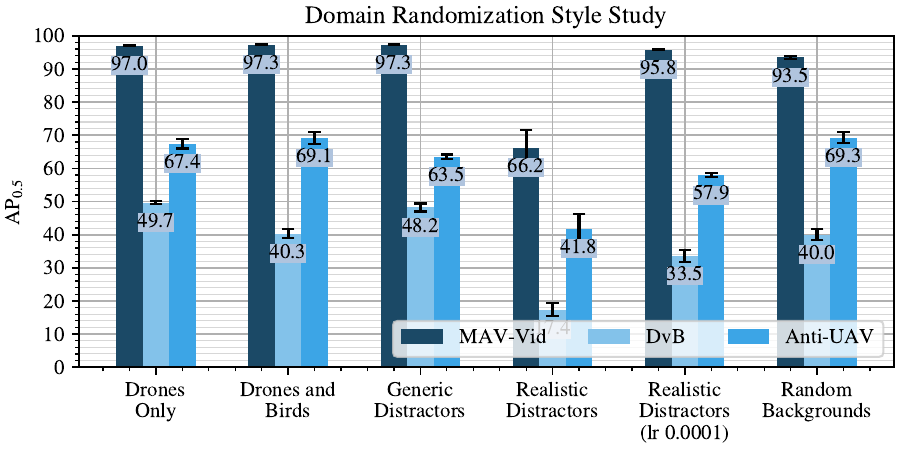}
\caption{Results of the domain randomization techniques study for drones only, drones and birds, drones with generic distractors, drones with realistic distractors, and drones with unrealistic backgrounds. Average Precision at .50 IoU ($AP_{0.5}$) on the y-axis, presented as a mean of 8 runs along with a 95 per cent confidence interval, and the selected domain randomization technique on the x-axis. Each bounding size is tested on the MAV-Vid, Drone-vs-Bird, and Anti-UAV datasets.  \label{fig:results:domain_randomization_study}}
\end{figure}  

Figure \ref{fig:results:domain_randomization_study} shows the results of the domain randomization style study. The drones only, drones and birds, and generic distractors datasets all perform reasonably well across the three test datasets with some small but noticeable differences. They all perform very similarly on the MAV-Vid dataset, achieving 97.0\% to 97.3\% AP. On the DvB dataset, the drones-only dataset achieves the highest $AP_{0.5}$ across the whole study, of 49.7\%. Generic distractors achieve a similar mean of 0.482. The drones and birds dataset surprisingly performs the worst on the DvB dataset out of the three, with a significant drop to 40.3\%. This result counters our hypothesis when designing this dataset - that adding synthetic bird distractors will improve the performance of the model, especially on the DvB dataset, which was designed with the intention of challenging detection networks to detect drones with birds as distractors. 
Overall, none of the domain randomization styles have achieved significant improvements over the drones-only dataset. 

\subsection{Comparison to Literature}\label{section:results:comparion_to_literature}
These experiments were designed based on the benchmark of Isaac-Medina et al. \cite{isaac-medinaUnmannedAerialVehicle2021}. 
Our experimental procedure slightly varies from Isaac-Medina et al. in that we present a mean and a standard deviation, while they present the results of a single run. 
Another difference is that Isaac-Medina et al. trained a separate model for each of the datasets. That means that for the MAV-Vid dataset, they split the dataset for training and evaluation. They trained a model on the training subset and then tested it on the evaluation subset. They then trained and evaluated another model from scratch for the DvB and Anti-UAV datasets. Hence, they present three sets of model weights, each one for the particular dataset. Instead, we present a single set of model weights, trained purely on the synthetic dataset, that has not seen any of the test datasets. Further, they tested a number of architectures together: Faster RCNN, SSD512, YOLOv3, and DETR. For consistency, we only compare to their Faster-RCNN results.

\begin{table*}\label{table:results:literature_comparison}
\caption{Comparison of our results, of the Faster-RCNN model trained on the drones-only dataset with 40 m bounds, to results from literature }
\centering
\small
\begin{tabularx}{\textwidth}{l|l|llllllllll}
\toprule
\textbf{Dataset} & \textbf{Model} & \textbf{AP} & \textbf{AP\textsubscript{0.5}} & \textbf{AP\textsubscript{0.75}} & \textbf{AP\textsubscript{S}} & \textbf{AP\textsubscript{M}} & \textbf{AP\textsubscript{L}} & \textbf{AR} & \textbf{AR\textsubscript{S}} & \textbf{AR\textsubscript{M}} & \textbf{AR\textsubscript{L}} \\
\midrule
MAV-VID & Ours & 0.526 & 0.970 & 0.503 & 0.058 & 0.493 & 0.586 & 0.599 & 0.402 & 0.580 & 0.664 \\
 & Isaac-Medina et al. \cite{isaac-medinaUnmannedAerialVehicle2021} & 0.592 & 0.978 & 0.672 & 0.154 & 0.541 & 0.656 & 0.659 & 0.369 & 0.621 & 0.721 \\
 \midrule
 Drone-vs-Bird & Ours & 0.187 & 0.498 & 0.075 & 0.170 & 0.336 & 0.294 & 0.246 & 0.307 & 0.496 & 0.322 \\
 & Isaac-Medina et al. & 0.283 & 0.632 & 0.197 & 0.218 & 0.473 & 0.506 & 0.356 & 0.298 & 0.546 & 0.512 \\
\midrule
 Anti-UAV-RGB & Ours & 0.299 & 0.678 & 0.173 & 0.000 & 0.304 & 0.291 & 0.359 & 0.000 & 0.449 & 0.343 \\
 & Isaac-Medina et al. & 0.581 & 0.977 & 0.641 & 0.523 & 0.623 & - & 0.636 & 0.602 & 0.663 & -  \\
 & Freudenmann et al. \cite{freudenmannExploitationDataAugmentation2021} & - & 0.542 & - & - & - & - & - & - & - & - \\
 & Freudenmann et al. & - & 0.787 & - & - & - & - & - & - & - & - \\
 & (With Augmentations)  &  &  &  &  &  &  &  &  &  & \\
 
\bottomrule
\label{table:results:literature_comparison}
\end{tabularx}
\end{table*}

Table \ref{table:results:literature_comparison} shows the comparison of the results of our Faster-RCNN model trained on the drones-only dataset (with 40 m bounds) compared with Isaac-Medina et al. Faster-RCNN benchmark results. We focus on the $AP_{0.5}$ metric because we believe it is the most important in this context. Because of imperfect labels in the real-world datasets, we don't believe the $AP$ and $AP_{0.75}$ to be as relevant. On the MAV-Vid test dataset, our model achieves a mean (across 8 runs) $AP_{0.5}$ of 0.970, compared with the 0.978 from Medina. On the DvB dataset, our model achieves a mean $AP_{0.5}$ of 0.498, compared with the 0.632 from Medina. This presents a considerable gap in performance of 13.4\%. Lastly, on the Anti-UAV dataset, our model achieves a mean $AP_{0.5}$ of 0.678, compared with the 0.977 from Medina. This is the biggest gap across the three test datasets of 29.9\%. Hence, our model translates very well to MAV-Vid, moderately to DvB, and poorly to Anti-UAV. 

We believe that the nature of the test datasets has a severe impact on how well our synthetic-data-trained model will translate to real life. MAV-Vid, the dataset on which our model performs well, contains videos of drones following other drones and ground-based cameras. The dataset is the most straightforward out of the three test datasets. DvB contains videos of drones filmed in complex environments. However, the majority of our HDRI backgrounds (that we used for generating the synthetic dataset) are skies. Our model performs reasonably well on the small and medium AP/recall (achieving 0.170 $AP_{S}$, and 0.336 $AP_{M}$ compared with 0.218 $AP_{S}$ and 0.473 $AP_{M}$ from Isaac-Medina et al.) but drops in performance on the large objects AP/recall (0.294 $AP_{L}$ compared with 0.506 $AP_{L}$ from Isaac-Medina et al.). The inclusion of more complex modelled environments as opposed to HDRIs only in our synthetic dataset could potentially improve the sim-to-real for this particular dataset. However, the realistic distractors dataset aimed at creating these realistic, complex distractors, did not improve the performance over the standard drones-only dataset. Lastly, the Anti-UAV dataset is unique in that it contains artefacts from the recording camera (including Chinese letters, numerical values, crosshairs, etc.). This makes models susceptible to false positives. We believe that the poor performance of our model on this dataset can be attributed to our model not being trained to handle these artefacts. Freudenmann et al. \cite{freudenmannExploitationDataAugmentation2021} performed a study on this, in which they found that a model trained on Drone vs Bird dataset gets an $AP_{0.5}$ of 54.2\% when tested on the Anti-UAV dataset. Note that we are unsure whether they used the same subset of the Anti-UAV dataset for evaluation, so the values should not be directly compared but can be used as guidance. They find that, by applying certain dataset augmentations they are able to increase the performance to 78.7\%. Because Isaac-Medina et al. trained their model on the Anti-UAV dataset, the model was trained not to be affected by these artefacts. By performing a data augmentation strategy or domain adaptation, we believe that the performance of our model could be increased. However, this falls outside the scope of this study. 
Finally, another aspect of the Anti-UAV dataset that has not been considered so far, is that a subset of the videos are filmed during nighttime. We included these in the test dataset to be consistent with Isaac-Medina et al., but as the synthetic dataset did not include any truly nighttime scenarios (where you truly cannot see the drone, and the only way to distinguish it is by its lights) it is not surprising that the performance drops during this particular comparison.

\section{Conclusion}\label{section:Conclusion}
To conclude, we have been successful at our original aim of creating a synthetic dataset using structured domain randomization and training a Faster-RCNN model which transfers to real-world datasets. We presented quantifiable results and compared them with real-world-based datasets. We have presented a Faster-RCNN model trained on a purely synthetic dataset that successfully bridges the sim-to-real gap on the real MAV-Vid dataset, achieving an $AP_{0.5}$ of 97.0\%, which is shy of the 97.8\% for an equivalent model trained on the MAV-Vid dataset. The model also performs reasonably on the Drone-vs-Bird (DvB) dataset, achieving $AP_{0.5}$ of 49.8\%, compared with 63.2\% for an equivalent model trained on the DvB dataset. The model translates poorly to the Anti-UAV dataset, achieving $AP_{0.5}$ of 67.8\%, compared with 97.7\% for an equivalent model trained on the DvB dataset. However, we believe this to be the case due to artefacts that are unique to the Anti-UAV dataset. The model still outperforms Freudenmann et al. \cite{freudenmannExploitationDataAugmentation2021} model which was trained on the Drone-vs-Bird dataset only, and achieved 54.2\% when tested on the Anti-UAV dataset. This shows that, despite not quite matching the benchmark, our weights prove to be generalizable. 

The purpose of this chapter is not to try and improve on the results of the Isaac-Medina et al. benchmark, but rather to examine the sim-to-real performance of a model trained on a purely synthetic dataset and understand the variables which impact sim-to-real transferability of the model. Hence, we did not try to find a model which performs well on a single dataset but rather aimed to investigate which variables affect the generalizability across all three of the real-life datasets. Our hypothesis is, that by improving the sim-to-real performance across the three datasets, we would also improve the performance for unseen datasets of drones. We have achieved this by focusing on the mean metrics as opposed to finding the well-performing outliers, which can be caused by the random seed which randomizes the convergence of NN models during training. 

Our domain randomization style study \ref{section:results:domain_randomization_styles} has failed to find evidence that any of the proposed styles significantly improve the sim-to-real transferability. This is surprising because it disputes our hypothesis based on the literature review that domain randomization and distractors improve the sim-to-real transferability. 
This could be explained by domain randomization not being applicable to the application of drone detection, our domain randomization styles not working for our test datasets, or our learning strategy favouring the drones-only dataset. Still, we conclude that we have failed to find evidence that the addition of distractors or randomization of backgrounds has improved the sim-to-real transferability. 

We believe this research to be an important step towards the use of synthetic data for creating operational neural network models for drone detection. We have shown the benefits or using synthetic data and the potential for generalizability, what works and what does not from general synthetic data literature, as well as the limitations such as inaccuracies within specific datasets.

\section{Further Work}\label{section:Further_Work}
Temporal networks - convolutional networks with multiple frames concatenated, or recurrent layers - are being to be explored in the literature (e.g. Thai et al. \cite{thaiSmallFlyingObject2023}). This is another area where synthetic data could be useful, although the flight mechanics of the drones should be modelled correctly. In this study, we focused on single-frame detection. Our dataset, as it stands, is also incompatible with this kind of architecture because the camera position changes every frame (so the animations are not consistent). However, this is something that could be modified to generate animations of drones flying realistic realistic trajectories. If modelled correctly, this presents a potential for much more accurate drone-vs-bird detection. In our results, we investigated the effects of JPEG compression, but the real-world datasets are also encoded using video compression which we were not able to test. Together with temporal networks, an animated dataset of drones and birds might be the answer to bridging the sim-to-real gap. 

We did not train our networks to detect the birds or distractor objects to be consistent with the test datasets. However, the datasets themselves do contain these in the segmentation masks, labelled in blue (drones are labelled in white). Detecting two classes might make the detector more accurate (although preliminary studies did not show this to be the case, hence it is not something we pursued with much detail). However, with the correct training strategy, it is something that might hold the potential for increasing the sim-to-real performance.

Because it was not the aim of our study, we did not consider modifying the default parameters of the Faster R-CNN network. However, an ablation study into the parameters of the Faster R-CNN network could be performed. This could include using a different pre-trained model, trying different learning strategies, freezing different layers, etc.

The presented segmentation masks can be used to train a segmentation model such as Mask-RCNN or U-Net to present pixel-accurate detections of drones. We did not do this in this study because it did not benefit the detection accuracy comparisons.

\section*{Acknowledgments}
This project was funded through the UK Government’s Industrial PhD Partnership (IPPs) Scheme by the Future Aviation Security Solutions Programme (FASS) - a joint Department for Transport and Home Office venture, in collaboration with Aveillant Ltd. and Autonomous Devices Ltd.

\bibliographystyle{IEEEtran}
\bibliography{MyLibrary.bib}

\begin{thebibliography}{10}
\providecommand{\url}[1]{#1}
\csname url@samestyle\endcsname
\providecommand{\newblock}{\relax}
\providecommand{\bibinfo}[2]{#2}
\providecommand{\BIBentrySTDinterwordspacing}{\spaceskip=0pt\relax}
\providecommand{\BIBentryALTinterwordstretchfactor}{4}
\providecommand{\BIBentryALTinterwordspacing}{\spaceskip=\fontdimen2\font plus
\BIBentryALTinterwordstretchfactor\fontdimen3\font minus \fontdimen4\font\relax}
\providecommand{\BIBforeignlanguage}[2]{{%
\expandafter\ifx\csname l@#1\endcsname\relax
\typeout{** WARNING: IEEEtran.bst: No hyphenation pattern has been}%
\typeout{** loaded for the language `#1'. Using the pattern for}%
\typeout{** the default language instead.}%
\else
\language=\csname l@#1\endcsname
\fi
#2}}
\providecommand{\BIBdecl}{\relax}
\BIBdecl

\bibitem{IncreaseUseDrones2023}
B.~News, ``Increase in use of drones for prison smuggling,'' \emph{BBC News}, Apr. 2023.

\bibitem{DrugsWeaponsSmuggled2022}
------, ``Drugs, weapons 'smuggled to prisoners by drone','' \emph{BBC News}, Feb. 2022.

\bibitem{HeathrowAirportDrone2019}
------, ``Heathrow airport: {{Drone}} sighting halts departures,'' \emph{BBC News}, Jan. 2019.

\bibitem{skopelitiFlightsDivertedEast2022}
C.~Skopeliti, ``Flights diverted at {{East Midlands}} airport after drone sightings,'' \emph{The Guardian}, Jun. 2022.

\bibitem{DublinAirportFlights2023}
``Dublin airport: {{Flights}} suspended for 30 minutes after drone sightings,'' https://www.bbc.com/news/articles/c72mvndez5jo, Feb. 2023.

\bibitem{UKCounterUnmannedAircraft}
``{{UK Counter-Unmanned Aircraft Strategy}},'' p.~38.

\bibitem{lykouDefendingAirportsUAS2020}
G.~Lykou, D.~Moustakas, and D.~Gritzalis, ``Defending {{Airports}} from {{UAS}}: {{A Survey}} on {{Cyber-Attacks}} and {{Counter-Drone Sensing Technologies}},'' \emph{Sensors}, vol.~20, no.~12, p. 3537, Jun. 2020.

\bibitem{krizhevskyImageNetClassificationDeep2017}
A.~Krizhevsky, I.~Sutskever, and G.~E. Hinton, ``{{ImageNet}} classification with deep convolutional neural networks,'' \emph{Communications of the ACM}, vol.~60, no.~6, pp. 84--90, May 2017.

\bibitem{redmonYouOnlyLook2016}
J.~Redmon, S.~Divvala, R.~Girshick, and A.~Farhadi, ``You {{Only Look Once}}: {{Unified}}, {{Real-Time Object Detection}},'' May 2016.

\bibitem{redmonYOLO9000BetterFaster2017}
J.~Redmon and A.~Farhadi, ``{{YOLO9000}}: {{Better}}, {{Faster}}, {{Stronger}},'' in \emph{2017 {{IEEE Conference}} on {{Computer Vision}} and {{Pattern Recognition}} ({{CVPR}})}.\hskip 1em plus 0.5em minus 0.4em\relax {Honolulu, HI}: {IEEE}, Jul. 2017, pp. 6517--6525.

\bibitem{liuSSDSingleShot2016}
W.~Liu, D.~Anguelov, D.~Erhan, C.~Szegedy, S.~Reed, C.-Y. Fu, and A.~C. Berg, ``{{SSD}}: {{Single Shot MultiBox Detector}},'' 2016, vol. 9905, pp. 21--37.

\bibitem{girshickFastRCNN2015}
R.~Girshick, ``Fast {{R-CNN}},'' Sep. 2015.

\bibitem{renFasterRCNNRealTime2016}
S.~Ren, K.~He, R.~Girshick, and J.~Sun, ``Faster {{R-CNN}}: {{Towards Real-Time Object Detection}} with {{Region Proposal Networks}},'' \emph{arXiv:1506.01497 [cs]}, Jan. 2016.

\bibitem{isaac-medinaUnmannedAerialVehicle2021}
B.~K.~S. {Isaac-Medina}, M.~Poyser, D.~Organisciak, C.~G. Willcocks, T.~P. Breckon, and H.~P.~H. Shum, ``Unmanned {{Aerial Vehicle Visual Detection}} and {{Tracking}} using {{Deep Neural Networks}}: {{A Performance Benchmark}},'' in \emph{2021 {{IEEE}}/{{CVF International Conference}} on {{Computer Vision Workshops}} ({{ICCVW}})}.\hskip 1em plus 0.5em minus 0.4em\relax {Montreal, BC, Canada}: {IEEE}, Oct. 2021, pp. 1223--1232.

\bibitem{vaswaniAttentionAllYou2017}
A.~Vaswani, N.~Shazeer, N.~Parmar, J.~Uszkoreit, L.~Jones, A.~N. Gomez, L.~Kaiser, and I.~Polosukhin, ``Attention {{Is All You Need}},'' Dec. 2017.

\bibitem{carionEndtoEndObjectDetection2020}
N.~Carion, F.~Massa, G.~Synnaeve, N.~Usunier, A.~Kirillov, and S.~Zagoruyko, ``End-to-{{End Object Detection}} with {{Transformers}},'' May 2020.

\bibitem{liuOutOfDistributionGeneralizationSurvey2023}
J.~Liu, Z.~Shen, Y.~He, X.~Zhang, R.~Xu, H.~Yu, and P.~Cui, ``Towards {{Out-Of-Distribution Generalization}}: {{A Survey}},'' Jul. 2023.

\bibitem{zhuangComprehensiveSurveyTransfer2020}
F.~Zhuang, Z.~Qi, K.~Duan, D.~Xi, Y.~Zhu, H.~Zhu, H.~Xiong, and Q.~He, ``A {{Comprehensive Survey}} on {{Transfer Learning}},'' Jun. 2020.

\bibitem{csurkaDomainAdaptationVisual2017}
G.~Csurka, ``Domain {{Adaptation}} for {{Visual Applications}}: {{A Comprehensive Survey}},'' Mar. 2017.

\bibitem{zhouDomainGeneralizationSurvey2022}
K.~Zhou, Z.~Liu, Y.~Qiao, T.~Xiang, and C.~C. Loy, ``Domain {{Generalization}}: {{A Survey}},'' \emph{IEEE Transactions on Pattern Analysis and Machine Intelligence}, pp. 1--20, 2022.

\bibitem{tobinDomainRandomizationTransferring2017}
J.~Tobin, R.~Fong, A.~Ray, J.~Schneider, W.~Zaremba, and P.~Abbeel, ``Domain randomization for transferring deep neural networks from simulation to the real world,'' in \emph{2017 {{IEEE}}/{{RSJ International Conference}} on {{Intelligent Robots}} and {{Systems}} ({{IROS}})}.\hskip 1em plus 0.5em minus 0.4em\relax {Vancouver, BC}: {IEEE}, Sep. 2017, pp. 23--30.

\bibitem{tremblayTrainingDeepNetworks2018}
J.~Tremblay, A.~Prakash, D.~Acuna, M.~Brophy, V.~Jampani, C.~Anil, T.~To, E.~Cameracci, S.~Boochoon, and S.~Birchfield, ``Training {{Deep Networks}} with {{Synthetic Data}}: {{Bridging}} the {{Reality Gap}} by {{Domain Randomization}},'' \emph{arXiv:1804.06516 [cs]}, Apr. 2018.

\bibitem{prakashStructuredDomainRandomization2019}
A.~Prakash, S.~Boochoon, M.~Brophy, D.~Acuna, E.~Cameracci, G.~State, O.~Shapira, and S.~Birchfield, ``Structured {{Domain Randomization}}: {{Bridging}} the {{Reality Gap}} by {{Context-Aware Synthetic Data}},'' in \emph{2019 {{International Conference}} on {{Robotics}} and {{Automation}} ({{ICRA}})}, May 2019, pp. 7249--7255.

\bibitem{borregoApplyingDomainRandomization2018}
J.~Borrego, A.~Dehban, R.~Figueiredo, P.~Moreno, A.~Bernardino, and J.~{Santos-Victor}, ``Applying {{Domain Randomization}} to {{Synthetic Data}} for {{Object Category Detection}},'' Jul. 2018.

\bibitem{alghonaimBenchmarkingDomainRandomisation2021}
R.~Alghonaim and E.~Johns, ``Benchmarking {{Domain Randomisation}} for {{Visual Sim-to-Real Transfer}},'' May 2021.

\bibitem{hinterstoisserPreTrainedImageFeatures2017}
S.~Hinterstoisser, V.~Lepetit, P.~Wohlhart, and K.~Konolige, ``On {{Pre-Trained Image Features}} and {{Synthetic Images}} for {{Deep Learning}},'' \emph{arXiv:1710.10710 [cs]}, Nov. 2017.

\bibitem{aydinDroneDetectionUsing2023}
B.~Aydin and S.~Singha, ``Drone {{Detection Using YOLOv5}},'' \emph{Eng}, vol.~4, no.~1, pp. 416--433, Mar. 2023.

\bibitem{samadzadeganDetectionRecognitionDrones2022a}
F.~Samadzadegan, F.~Dadrass~Javan, F.~Ashtari~Mahini, and M.~Gholamshahi, ``Detection and {{Recognition}} of {{Drones Based}} on a {{Deep Convolutional Neural Network Using Visible Imagery}},'' \emph{Aerospace}, vol.~9, no.~1, p.~31, Jan. 2022.

\bibitem{fanObjectDetectionAlgorithm2022a}
Y.~Fan, O.~Li, and G.~Liu, ``An {{Object Detection Algorithm}} for {{Rotary-Wing UAV Based}} on {{AWin Transformer}},'' \emph{IEEE Access}, vol.~10, pp. 13\,139--13\,150, 2022.

\bibitem{dadrassjavanModifiedYOLOv4Deep2022}
F.~Dadrass~Javan, F.~Samadzadegan, M.~Gholamshahi, and F.~Ashatari~Mahini, ``A {{Modified YOLOv4 Deep Learning Network}} for {{Vision-Based UAV Recognition}},'' \emph{Drones}, vol.~6, no.~7, p. 160, Jul. 2022.

\bibitem{freudenmannExploitationDataAugmentation2021}
L.~Freudenmann, L.~Sommer, and A.~Schumann, ``Exploitation of data augmentation strategies for improved {{UAV}} detection,'' in \emph{Automatic {{Target Recognition XXXI}}}, vol. 11729.\hskip 1em plus 0.5em minus 0.4em\relax {SPIE}, Apr. 2021, pp. 119--132.

\bibitem{mediavillaDetectingAerialObjects2021}
C.~Mediavilla, L.~Nans, D.~Marez, and S.~Parameswaran, ``Detecting aerial objects: Drones, birds, and helicopters,'' in \emph{Artificial {{Intelligence}} and {{Machine Learning}} in {{Defense Applications III}}}, J.~Dijk, Ed.\hskip 1em plus 0.5em minus 0.4em\relax {Online Only, Spain}: {SPIE}, Sep. 2021, p.~18.

\bibitem{thaiSmallFlyingObject2023}
P.~Thai, S.~Alam, N.~Lilith, and B.~Nguyen, ``Small {{Flying Object Detection}} and {{Tracking}} in {{Digital Airport Tower}} through {{Spatial-Temporal ConvNets}},'' {In Review}, Preprint, Mar. 2023.

\bibitem{crayeSpatioTemporalSemanticSegmentation2019}
C.~Craye and S.~Ardjoune, ``Spatio-{{Temporal Semantic Segmentation}} for {{Drone Detection}},'' in \emph{2019 16th {{IEEE International Conference}} on {{Advanced Video}} and {{Signal Based Surveillance}} ({{AVSS}})}.\hskip 1em plus 0.5em minus 0.4em\relax {Taipei, Taiwan}: {IEEE}, Sep. 2019, pp. 1--5.

\bibitem{sangamTransVisDroneSpatioTemporalTransformer2023}
T.~Sangam, I.~R. Dave, W.~Sultani, and M.~Shah, ``{{TransVisDrone}}: {{Spatio-Temporal Transformer}} for {{Vision-based Drone-to-Drone Detection}} in {{Aerial Videos}},'' Aug. 2023.

\bibitem{rozantsevRenderingSyntheticImages2015}
A.~Rozantsev, V.~Lepetit, and P.~Fua, ``On {{Rendering Synthetic Images}} for {{Training}} an {{Object Detector}},'' \emph{Computer Vision and Image Understanding}, vol. 137, pp. 24--37, Aug. 2015.

\bibitem{marezUAVDetectionDataset2020}
D.~Marez, S.~Borden, and L.~Nans, ``{{UAV}} detection with a dataset augmented by domain randomization,'' in \emph{Geospatial {{Informatics X}}}, vol. 11398.\hskip 1em plus 0.5em minus 0.4em\relax {SPIE}, May 2020, pp. 39--50.

\bibitem{pengUsingImagesRendered2018}
J.~Peng, C.~Zheng, T.~Cui, Y.~Cheng, and L.~Si, ``Using {{Images Rendered}} by {{PBRT}} to {{Train Faster R-CNN}} for {{UAV Detection}},'' in \emph{26. {{International Conference}} in {{Central Europe}} on {{Computer Graphics}}, {{Visualization}} and {{Computer Vision}}'2017}, 2018.

\bibitem{dieterQuantifyingSimulationReality2023}
T.~R. Dieter, A.~Weinmann, S.~J{\"a}ger, and E.~Brucherseifer, ``Quantifying the {{Simulation}}{\textendash}{{Reality Gap}} for {{Deep Learning-Based Drone Detection}},'' \emph{Electronics}, vol.~12, no.~10, p. 2197, Jan. 2023.

\bibitem{scholesDronePoseIdentificationSegmentation2021}
S.~Scholes, A.~Ruget, G.~{Mora-Martin}, F.~Zhu, I.~Gyongy, and J.~Leach, ``{{DronePose}}: {{The}} identification, segmentation, and orientation detection of drones via neural networks,'' \emph{arXiv:2112.05488 [cs]}, Dec. 2021.

\bibitem{liScarceDataDriven2021}
C.~Li, S.~C. Sun, Z.~Wei, A.~Tsourdos, and W.~Guo, ``Scarce {{Data Driven Deep Learning}} of {{Drones}} via {{Generalized Data Distribution Space}},'' \emph{arXiv:2108.08244 [cs]}, Aug. 2021.

\bibitem{carrioDroneDetectionUsing2018}
A.~Carrio, S.~Vemprala, A.~Ripoll, S.~Saripalli, and P.~Campoy, ``Drone {{Detection Using Depth Maps}},'' Aug. 2018.

\bibitem{wisniewskiDroneModelClassification2022}
M.~Wisniewski, Z.~A. Rana, and I.~Petrunin, ``Drone {{Model Classification Using Convolutional Neural Network Trained}} on {{Synthetic Data}},'' p.~20, 2022.

\bibitem{wisniewskiDroneModelIdentification2021}
------, ``Drone {{Model Identification}} by {{Convolutional Neural Network}} from {{Video Stream}},'' in \emph{2021 {{IEEE}}/{{AIAA}} 40th {{Digital Avionics Systems Conference}} ({{DASC}})}.\hskip 1em plus 0.5em minus 0.4em\relax {San Antonio, TX, USA}: {IEEE}, Oct. 2021, pp. 1--8.

\bibitem{blenderfoundationBlender3DModelling}
B.~Foundation, ``Blender - a {{3D}} modelling and rendering package.''

\bibitem{linMicrosoftCOCOCommon2015}
T.-Y. Lin, M.~Maire, S.~Belongie, L.~Bourdev, R.~Girshick, J.~Hays, P.~Perona, D.~Ramanan, C.~L. Zitnick, and P.~Doll{\'a}r, ``Microsoft {{COCO}}: {{Common Objects}} in {{Context}},'' Feb. 2015.

\bibitem{paszkePyTorchImperativeStyle}
A.~Paszke, S.~Gross, F.~Massa, A.~Lerer, J.~Bradbury, G.~Chanan, T.~Killeen, Z.~Lin, N.~Gimelshein, L.~Antiga, A.~Desmaison, A.~Kopf, E.~Yang, Z.~DeVito, M.~Raison, A.~Tejani, S.~Chilamkurthy, B.~Steiner, L.~Fang, J.~Bai, and S.~Chintala, ``{{PyTorch}}: {{An Imperative Style}}, {{High-Performance Deep Learning Library}},'' p.~12.

\bibitem{VisionTorchvisionModels}
``Vision/torchvision/models/detection/faster\_rcnn.py at main {$\cdot$} pytorch/vision,'' https://github.com/pytorch/vision/blob/main/ torchvision/models/detection/faster\_rcnn.py.

\bibitem{poyserImpactLossyImage2020}
M.~Poyser, A.~{Atapour-Abarghouei}, and T.~P. Breckon, ``On the {{Impact}} of {{Lossy Image}} and {{Video Compression}} on the {{Performance}} of {{Deep Convolutional Neural Network Architectures}},'' Jul. 2020.

\bibitem{jungImgaug2023}
A.~Jung, ``Imgaug,'' Nov. 2023.

\bibitem{rodriguez-ramosAdaptiveInattentionalFramework2020}
A.~{Rodriguez-Ramos}, J.~{Rodriguez-Vazquez}, C.~Sampedro, and P.~Campoy, ``Adaptive {{Inattentional Framework}} for {{Video Object Detection With Reward-Conditional Training}},'' \emph{IEEE Access}, vol.~8, pp. 124\,451--124\,466, 2020.

\bibitem{colucciaDronevsBirdDetectionChallenge}
A.~Coluccia, A.~Fascista, A.~Schumann, L.~Sommer, M.~Ghenescu, A.~O. Avenue, and T.~Piatrik, ``Drone-vs-{{Bird Detection Challenge}} at {{IEEE AVSS2019}},'' p.~7.

\bibitem{colucciaDroneVsBird2021}
A.~Coluccia, A.~Fascista, A.~Schumann, L.~Sommer, A.~Dimou, D.~Zarpalas, M.~M{\'e}ndez, D.~{de la Iglesia}, I.~Gonz{\'a}lez, J.-P. Mercier, G.~Gagn{\'e}, A.~Mitra, and S.~Rajashekar, ``Drone vs. {{Bird Detection}}: {{Deep Learning Algorithms}} and {{Results}} from a {{Grand Challenge}},'' \emph{Sensors}, vol.~21, no.~8, p. 2824, Jan. 2021.

\bibitem{jiangAntiUAVLargeMultiModal2021}
N.~Jiang, K.~Wang, X.~Peng, X.~Yu, Q.~Wang, J.~Xing, G.~Li, J.~Zhao, G.~Guo, and Z.~Han, ``Anti-{{UAV}}: {{A Large Multi-Modal Benchmark}} for {{UAV Tracking}},'' \emph{arXiv:2101.08466 [cs]}, Feb. 2021.

\bibitem{picardTorchManual_seed34072023}
D.~Picard, ``Torch.manual\_seed(3407) is all you need: {{On}} the influence of random seeds in deep learning architectures for computer vision,'' May 2023.

\bibitem{taleb2007black}
N.~N. Taleb, \emph{The Black Swan: {{The}} Impact of the Highly Improbable}.\hskip 1em plus 0.5em minus 0.4em\relax {Random house}, 2007, vol.~2.

\end{thebibliography}
\newpage

\end{document}